\documentclass{article}

\usepackage{microtype}
\usepackage{graphicx}
\usepackage{subfigure}
\usepackage{booktabs} 

\usepackage{hyperref}

\usepackage[accepted]{icml2025}

\usepackage{amsmath}
\usepackage{amssymb}
\usepackage{mathtools}
\usepackage{amsthm}

\usepackage[capitalize,noabbrev]{cleveref}

\newcommand{\revision}[1]{\textcolor{black}{#1}}

\theoremstyle{plain}

\theoremstyle{definition}

\theoremstyle{remark}

\usepackage[textsize=tiny]{todonotes}

\PassOptionsToPackage{svgnames}{xcolor}
\usepackage{xcolor}
\usepackage{xspace}
\usepackage{tcolorbox}
\usepackage{multicol}
\usepackage{algorithmic}
\usepackage{graphicx}
\usepackage{array}
\usepackage[skip=6pt]{caption}

\newcommand{\algname}{\textsc{GRAPE}\xspace}

\newcommand{\alg}{{TPO}\xspace}

\usepackage{dialogue}
\tcbuselibrary{breakable}
\tcbuselibrary{skins}
\newtcolorbox{boxone}{
  breakable=true,
  colback=blue!10,
  boxrule=0.5mm,
  colframe=blue!30!black,
  fonttitle=\bfseries,
  title=Prompt Template for Multi-stage Cost Proposal,
}

\newtcolorbox{boxtwo}{
  breakable=true,
  colback=blue!10,
  boxrule=0.5mm,
  colframe=blue!30!black,
  fonttitle=\bfseries,
  title=Cost Functions for Cost-Efficiency Alignment,
}

\newtcolorbox{boxthree}{
  breakable=true,
  colback=blue!10,
  boxrule=0.5mm,
  colframe=blue!30!black,
  fonttitle=\bfseries,
  title=Cost Functions for Safety Alignment,
}
\newtcolorbox{boxfour}{
  breakable=true,
  colback=blue!10,
  boxrule=0.5mm,
  colframe=blue!30!black,
  fonttitle=\bfseries,
  title=Cost Functions for Task Completion Alignment,
}

\newcommand{\figref}[1]{Figure~\ref{#1}}
\newcommand{\tabref}[1]{Table~\ref{#1}}
\newcommand{\eqnref}[1]{\text{Eq.}~(\ref{#1})}

\newcommand{\algoref}[1]{Algorithm~\ref{#1}}

\usepackage{multirow}
\icmltitlerunning{\algname: Generalizing Robot Policy via Preference Alignment}

\begin{document}

\twocolumn[
\icmltitle{\algname: Generalizing Robot Policy via Preference Alignment}



\icmlsetsymbol{equal}{*}

\begin{icmlauthorlist}
\icmlauthor{Zijian Zhang}{equal,unc}
\icmlauthor{Kaiyuan Zheng}{equal,uw}
\icmlauthor{Zhaorun Chen}{equal,uchicago}
\icmlauthor{Joel Jang}{uw}
\icmlauthor{Yi Li}{uw}
\icmlauthor{Siwei Han}{unc}\\
\icmlauthor{Chaoqi Wang}{uchicago}
\icmlauthor{Mingyu Ding}{unc}
\icmlauthor{Dieter Fox}{uw}
\icmlauthor{Huaxiu Yao}{unc}
\end{icmlauthorlist}

\icmlaffiliation{unc}{UNC-Chapel Hill, Chapel Hill, NC, USA}
\icmlaffiliation{uw}{University of Washington, Seattle, WA, USA}
\icmlaffiliation{uchicago}{University of Chicago, Chicago, IL, USA}

\icmlcorrespondingauthor{Huaxiu Yao}{huaxiu@cs.unc.edu}

\icmlkeywords{Machine Learning, ICML}

\vskip 0.3in
]



\printAffiliationsAndNotice{\icmlEqualContribution} 

\begin{abstract}
Despite the recent advancements of vision-language-action (VLA) models on a variety of robotics tasks, they suffer from critical issues such as poor generalizability to unseen tasks, due to their reliance on behavior cloning exclusively from successful rollouts. Furthermore, they are typically fine-tuned to replicate demonstrations collected by experts under different settings, thus introducing distribution bias and limiting their adaptability to diverse manipulation objectives, such as efficiency, safety, and task completion. To bridge this gap, we introduce \textbf{\algname}: \textbf{G}ene\textbf{ra}lizing Robot Policy via \textbf{P}r\textbf{e}ference Alignment.  Specifically, \algname aligns VLAs on a trajectory level and implicitly models reward from both successful and failure trials to boost generalizability to diverse tasks. Moreover, \algname breaks down complex manipulation tasks to independent stages and automatically guides preference modeling through customized spatiotemporal constraints with keypoints proposed by a large vision-language model. Notably, these constraints are flexible and can be customized to align the model with varying objectives, such as safety, efficiency, or task success. We evaluate \algname across a diverse array of tasks in both real-world and simulated environments. Experimental results demonstrate that \algname enhances the performance of state-of-the-art VLA models, increasing success rates on in-domain and unseen manipulation tasks by 51.79\% and 58.20\%, respectively. Additionally, \algname can be aligned with various objectives, such as safety and efficiency, reducing collision rates by 37.44\% and rollout step-length by 11.15\%, respectively.
%


\end{abstract}    
\section{Introduction}
\label{sec:intro}

The recent rapid proliferation of vision-language-action (VLA) models has streamlined general robotic manipulation tasks, demonstrating impressive capability across a range of tasks under controlled environmental variations~\cite{black2024pi_0, kim2024openvla, team2024octo, brohan2023rt}. However, these models face several critical challenges such as poor generalizability across new environments, objects, tasks, and semantic contexts~\cite{kim2024openvla}. A significant factor contributing to this limitation is their reliance on \textit{supervised fine-tuning} (SFT), where VLAs simply imitate actions from successful rollouts via behavior cloning while not developing a holistic understanding of the task goal or potential failure patterns~\cite{kumar2021should}. While reinforcement learning (RL) algorithms such as PPO~\cite{schulman2017proximal} have proved promising in enhancing their generalizability~\cite{zhai2024fine}, the high cost of gathering sufficient online trajectories and explicitly defining reward make them impractical for training VLA~\cite{team2024octo}.

\begin{figure}[t!]
    \centering
    \includegraphics[width=0.4\textwidth]{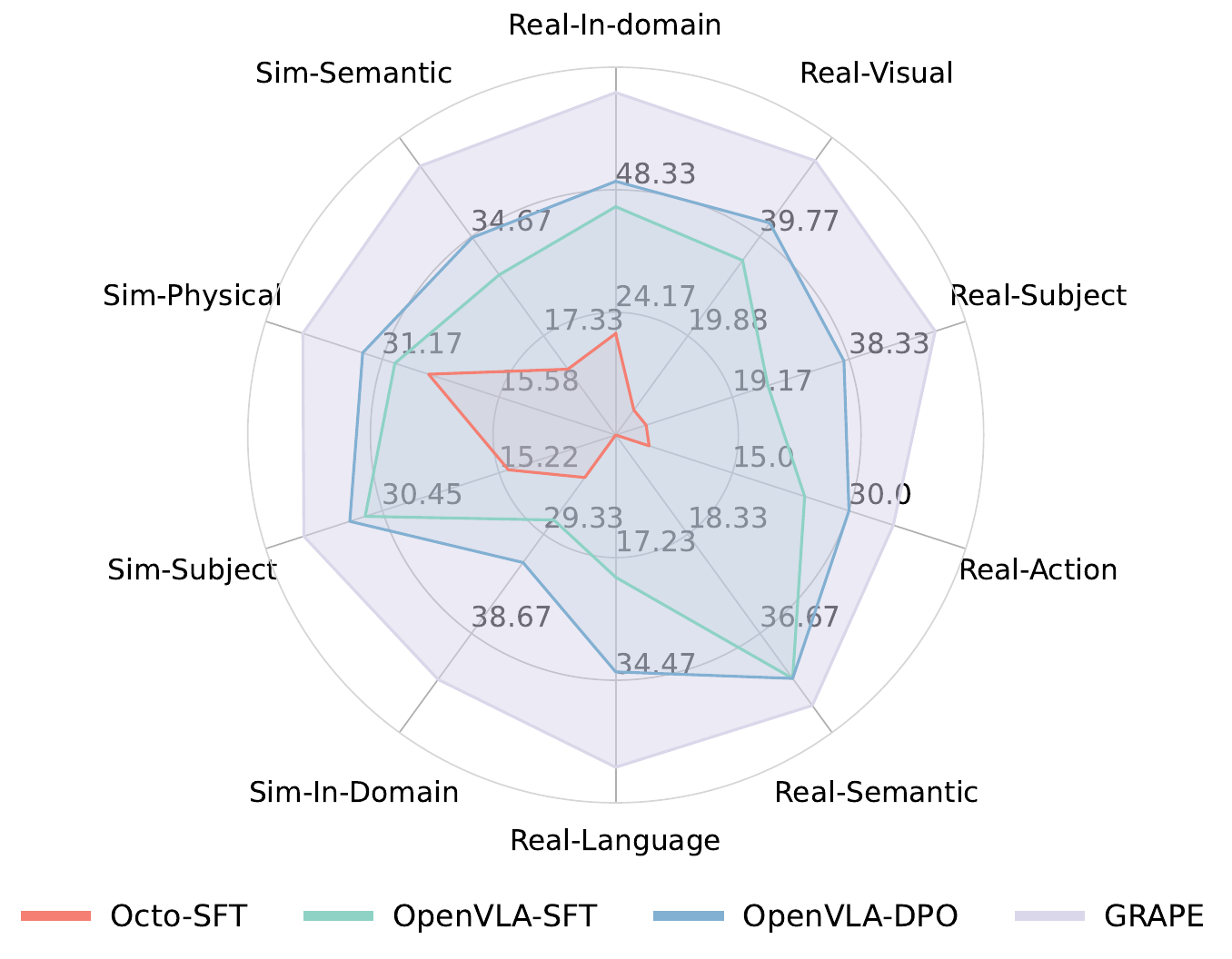}
    \caption{
    Comparison of \algname with SOTA VLA models fine-tuned on the same data across a large variety of generalization and in-domain tasks in both real-world and simulated environments.
}
\label{fig:overview}
\end{figure}

\begin{figure*}[ht!]
    \centering
    \includegraphics[width=1.0\textwidth]{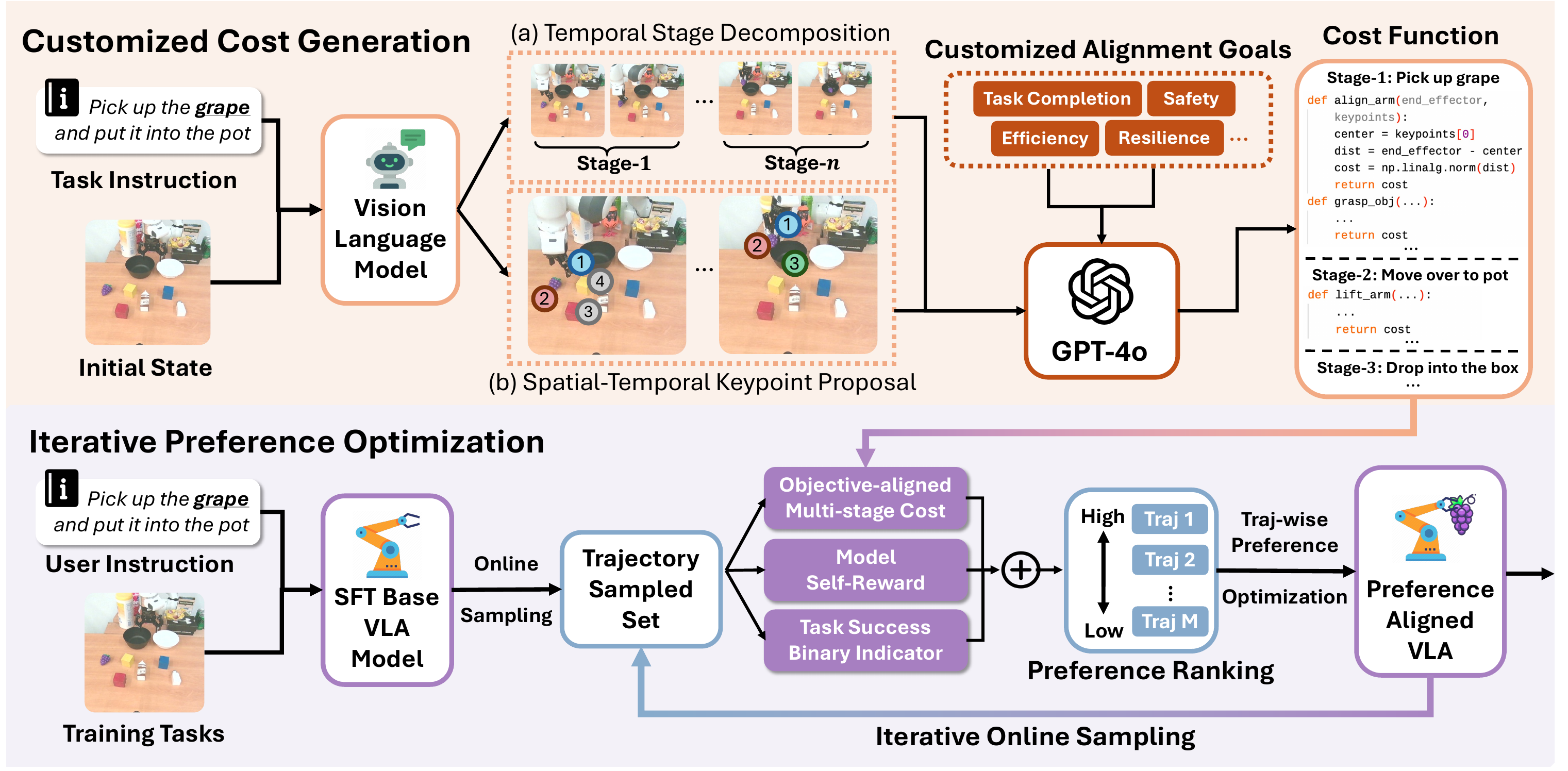}
    \caption{
    \textbf{Overview of \algname.} \algname first uses a VLM to decompose a manipulation task (\textbf{top}) into temporal stages and identify key spatial points for each subtask. Given user-specified alignment goals, it prompts a VLM to generate cost functions for each stage. During iterative preference optimization (\textbf{bottom}), offline trajectories are sampled from the base VLA model, scored using multi-stage cost, self-evaluation and task success indicators, and ranked to form preferences. \algname then optimizes the VLA models iteratively until convergence.
}
\label{fig:algorithm}
\vspace{-1.5em}
\end{figure*}
Furthermore, training VLAs to solely replicate expert behaviors often results in \textit{behavior collapse}~\cite{kumar2024training} where the planned trajectories are often suboptimal~\cite{kim2024openvla}. This is because the SFT datasets are usually uncurated and consist of offline demonstrations collected from experts that embed implicitly different values (e.g. task completion, safety, and cost-efficiency) that are not clearly defined within the data~\cite{o2023open, walke2023bridgedata}. Simply imitating these behaviors via SFT can potentially confuse the model and result in suboptimal trajectories that deviate from the actual objective of the demonstrations.
Some approaches attempt to address this challenge by explicitly defining a set of objectives and solving them hierarchically~\cite{huang2024rekep}. However, this approach incurs additional inference overhead and lacks scalability~\cite{li2024hamster}.

To address these issues, we propose \textbf{\algname}: \textbf{G}ene\textbf{ra}lizing Robot Policy via \textbf{P}r\textbf{e}ference Alignment to alleviate the high cost of training VLAs with RL objective, while offering flexibility for aligning towards customized manipulation objectives. 
As shown in~\figref{fig:algorithm}, \algname introduces \textit{trajectory-wise preference optimization} (TPO) to align VLA policies on a trajectory level by implicitly modeling reward from both successful and failure trials, boosting generalizability to diverse tasks. 
To further alleviate the difficulty in ranking trajectories and providing preferences towards arbitrary alignment objectives, \algname proposes to decompose the complex manipulation tasks into multiple independent stages and adopt a large vision model to propose keypoints for each stage, each associated with a spatial-temporal constraint.
Notably, these constraints are flexible and can be customized to align the model with varying manipulation objectives, such as task completion, robot-interaction safety, and cost-efficiency.
%
%
We evaluate \algname across a wide range of real-world tasks and two simulated environments. 
Experimental results show that \algname outperforms state-of-the-art VLA models, improving success rates on both in-domain and unseen manipulation tasks by 51.79\% and 58.20\%, respectively. Moreover, \algname can be aligned to diverse objectives such as safety and efficiency, to further reduce collision rate by 37.44\% and rollout step-length by 11.15\%, respectively.

\section{Generalizing Robot Policy via Preference Alignment}

\subsection{Preliminaries}

During inference, a VLA typically initializes with an task instruction $q$, and at each timestep $t$, it takes an environment observation $o_t$ (usually an image) and outputs an action $a_t$, where we can denote $\pi_{\theta}(a_i | (o_i, q))$ as the action policy of a VLA parameterized by $\theta$. To complete the task, VLA iteratively interacts with the environment and obtains a trajectory $\zeta = \{o_1, a_1, \cdots, o_T, a_T | q\}$ of length $T$. Typically, VLAs are fine-tuned to imitate expert behaviors via SFT:

%
\begin{equation}\label{eqn:sft}
\small
\mathcal{L}_{\text{SFT}} = - \sum_{(\zeta, q) \in \mathcal{D}} \sum_{t=1}^{T} \log p(a_t | o_t, q; \pi_{\theta}),
\end{equation}
where $\mathcal{D} = \{(\zeta_1, q_1), \dots, (\zeta_N, q_N)\}$ denotes the training set containing $N$ expert trajectories. Specifically, $\mathcal{L}_{\text{SFT}}$ enforces VLA to memorize the action associated with each observation sampled from a distribution $\mathbb{P}_\mathcal{D}$, resulting in poor generalizability to new task settings.
It is worth to note that while we follow~\citet{o2023open, brohan2023rt} and consider the step-wise policy based on the Markov decision process (MDP) assumption~\cite{sutton2018reinforcement}, our approach can be easily adapted to both non-MDP case which takes past interaction histories (usually a video or a series of images) as state~\cite{cheang2024gr} and diffusion policy~\cite{chi2023diffusion} which generates multiple future steps all at once~\cite{team2024octo}.

\subsection{\alg: Trajectory-wise Preference Optimization}

To improve generalization, we follow~\citet{schulman2017proximal, bai2022training} and further fine-tune VLA policies via RL objective. Let $r_{\phi}$ denote a reward function parameterized by $\phi$, we have
\begin{equation}
\small
\label{eqn:rl}
\max_{\pi_{\theta}} \mathbb{E}_{\zeta \sim \pi_\theta} \left[ r_{\phi}(\zeta) \right] - \beta D_{\text{KL}} \left[ \pi_{\theta}(\zeta) \parallel \pi_{\text{ref}}(\zeta) \right],
\end{equation}
where $\beta$ controls the deviation from the base reference policy $\pi_{\text{ref}}$ trained via SFT in~\eqnref{eqn:sft} and $\pi(\zeta, q)$ is the likelihood of policy $\pi$ generating the entire trajectory $\zeta$ under instruction $q$.
Then we follow~\citet{rafailov2024direct} and derive the analytical reparameterization of the trajectory reward $r(\zeta)$ as:
\begin{equation}
\small
\label{eqn:reward}
    r(\zeta, q) = \beta \log\frac{\pi_\theta(\zeta \mid q)}{\pi_{\text{ref}}(\zeta \mid q)} + \beta \log Z(\zeta).
\end{equation}
Similar to~\citet{rafailov2024direct}, we adopt the Bradley-Terry (BT)~\cite{bradley1952rank} model and model $r_{\phi}$ from a set of trajectories ranked with preferences. Specifically, let $\zeta_w$ and $\zeta_l$ denotes the chosen and rejected trajectory starting from the same initial state, we can formulate the trajectory-wise reward modeling objective as:
%
%
\begin{equation}
\small
\label{eqn:bt_model}
P\left( \zeta_w \succ \zeta_l \right) = \frac{\exp\left( r(\zeta_w), q \right)}{\exp\left( r(\zeta_w), q \right) + \exp\left( r(\zeta_l), q \right)}.
\end{equation}
Then, we follow~\citet{rafailov2024direct} and substitute~\eqnref{eqn:reward} into~\eqnref{eqn:bt_model} and obtain the following \textit{trajectory-wise preference optimization} (TPO) loss $\mathcal{L}_{\text{TPO}}$ equivalent to~\eqnref{eqn:rl}:

{\small
\begin{equation}
\small
\label{eqn:tpo}
\mathcal{L}_{\text{TPO}}=-\mathbb{E}_{(\zeta_w, \zeta_l) \sim \mathcal{D}} \left[ \log \sigma \left( \beta \left( \log \frac{\pi_\theta(\zeta_w)}{\pi_{\text{ref}}(\zeta_w)} - \log \frac{\pi_\theta(\zeta_l)}{\pi_{\text{ref}}(\zeta_l)} \right) \right) \right],
\end{equation}}

\noindent where we can further draw from MDP and decompose the likelihood of a trajectory $\zeta$ into individual state-action pairs, i.e., \begin{small}$\pi(\zeta, q) = \prod_{i=1}^T \pi(a_i \mid(o_i, q))$\end{small} and further obtain
\begin{equation}
\small
\label{eqn:trajectory_log_likelihood_ratio}
\log \frac{\pi_\theta(\zeta, q)}{\pi_{\text{ref}}(\zeta, q)} = \sum_{t=1}^{T} \log \frac{\pi_\theta(a_i \mid(o_i, q))}{\pi_{\text{ref}}(a_i \mid(o_i, q))}.
\end{equation}
Then we can substitute \eqnref{eqn:trajectory_log_likelihood_ratio} into \eqnref{eqn:tpo} to obtain the TPO loss $\mathcal{L}_{\text{TPO}}$ in terms of step-wise state-action pairs.
Our TPO loss~\eqnref{eqn:trajectory_log_likelihood_ratio} is beneficial as it: (1) aligns policy $\pi_\theta$ globally towards human preferences on a trajectory level while simply using step-wise rollouts collected by VLAs; (2) it stabilizes the policy and steers it towards the final goal by backpropagating the gradients throughout all the state-action pairs along the trajectory; (3) it significantly boosts generalizability by learning from both successful and failed trajectories via a RL objective.
Although \citet{finn2016guided} indicates that expanding the size of the sampled trajectory can reduce the bias in reward modeling, it also increases the training costs. Thus while our method can be easily scaled up, we keep our discussion to the binary case where only one chosen/rejected trajectory is present.

\subsection{Guided-Cost Preference Generation}

While given the TPO objective~\eqnref{eqn:tpo} we can align the policy towards arbitrary objectives defined through trajectories ranked by the corresponding preference, it incurs high costs as it requires human expertise and lengthy manual annotation. Thus to better scale up the preference synthesis towards arbitrary alignment objectives (e.g. task completion, safety, efficiency), we propose \textit{Guided-Cost Preference Generation (GCPG)} to automatically curate such preferences that integrate different alignment objectives.

\subsubsection{Multi-Stage Temporal Keypoint Constraints}


Building on insights from~\citet{huang2024rekep}, we address the complexity of specifying precise trajectory preferences for complex manipulation tasks by decomposing trajectories into temporal stages and assigning costs to quantify performance at each stage. Then, we aggregate these stage-specific costs to obtain a holistic evaluation for each trajectory.
Specifically, we adopt a VLM-based stage decomposer $\mathcal{M}_D$ (detailed in Appendix~\ref{sec:app_description}), to partition a trajectory $\zeta$ into a sequence of $\mathbf{S}$ consecutive stages, formulated as
\begin{equation}
\small
\label{eqn:decomposition}
    \{\zeta^1, \dots, \zeta^S\} = \mathcal{M}_D(\zeta, q), \quad \zeta^i = \{(o_t^i, a_t^i)\}_{t=1}^{T_i},
\end{equation}
where $\zeta^i$ represents the $i^\text{th}$ stage of trajectory $\zeta$. 

After obtaining the stage decomposition, we further employ a vision-language model (e.g. DINOv2~\cite{oquab2023dinov2}) to identify keypoints that serve as reference metrics across each stage. Then we prompt a powerful LLM~\cite{achiam2023gpt} to propose cost functions (see examples in Appendix ~\ref{sec:case_study_cost}.) for each stage that corresponds with the alignment objective, where lower cost indicates better objective compliance. 
Specifically, the cost $C^{S_i}(\{\kappa_{S_i}\})$  at stage $S_i$ is calculated using its corresponding keypoints $\{\kappa_{S_i}\}$.
%
%
%
%
%
%
%

Then to aggregate the costs for the entire trajectory, instead of summing each stage linearly, we apply an exponential decay to capture the casual dependencies of each temporal stage (e.g. if a trajectory incurs high costs in preceding stages it is not expected to perform well subsequently), defined as the \textit{external reward}:
%
\begin{equation}
\small
\label{eqn:overall_cost}
R_{\text{ext}}(\zeta) = \prod_{i=1}^{\mathbf{S}} e^{-C^{S_i}(\{\kappa_{S_i}\})}
\end{equation}
\noindent where~\eqnref{eqn:overall_cost} aggregates the individual costs and sub-objectives from each stage to tackle the curse of dimensionality and effectively adhere to the customized alignment.



\subsubsection{Guided-Cost Preference Generation}

To further improve the stability and optimality of the preference synthesis, we draw inspirations from self-rewarding~\cite{zhou2024calibrated} and determine that \textit{a more optimal trajectory should be confirmed by both the external judge (as in~\eqnref{eqn:overall_cost}) and the model itself}. Thus we incorporate two additional rewards and obtain the GCPG reward:
\begin{equation}
\small
\label{eqn:overall}
R_{\text{GCPG}}(\zeta) = \lambda_1 R_\text{self}(\zeta) +  \lambda_2 R_\text{ext}(\zeta) +  \lambda_3 I_{\text{success}}(\zeta)
\end{equation}
where $R_\text{self}(\zeta)$ is the self-evaluated score provided by $\pi$, which equals the log-likelihood of generating trajectory $\zeta$: 
\begin{equation}
\small
\label{eqn:self_traj}
R_\text{self}(\zeta) =\log(\pi(\zeta, q)) = \log(\prod_{i=1}^T\pi(a_i \mid(o_i, q)))
\end{equation}
and $I_{\text{success}}(\zeta)$ is a binary indicator function that indicates whether the trajectory $\zeta$ successfully completes the task:
\begin{equation}
\small
\label{eqn:indicator}
I_{\text{success}}(\zeta) =
\begin{cases} 
1, & \text{if } \zeta \text{ is successful}, \\
0, & \text{otherwise}.
\end{cases}
\end{equation}
where $\lambda$ are the weight parameters that adjust the importance of each reward. Intuitively, \eqnref{eqn:self_traj} can be seen as a dense approximation of the sparse signal provided by~\eqnref{eqn:indicator}, which are further calibrated by~\eqnref{eqn:overall_cost} to obtain a holistic evaluation of the trajectory that accounts for both its optimality and degree of alignment to a customized objective specified through the external reward in~\eqnref{eqn:overall_cost}.

\subsection{Iterative Preference Optimization}

After generating the preference, we then discuss our iterative preference optimization strategy. Inspired by the practices of on-policy RL~\cite{schulman2017proximal} which often yield more optimal policy than off-policy training, we iteratively fine-tune the SFT VLA model via TPO with trajectories collected online. For example, during the $k^\text{th}$ iteration, we (1) first sample numerous trajectories for a variety of tasks and obtain $\mathcal{D}^k$; (2) then we calculate the costs for each trajectory using~\eqnref{eqn:overall} and rank these trajectories accordingly per task; (3) we pair the top-$m$ and bottom-$m$ trajectories with each other for each task, and obtain $m^2$ chosen-rejected trajectory pairs; (4) then we fine-tune the same sampling policy with TPO via~\eqnref{eqn:tpo} and obtain an updated policy. We iterate this process for $K$ times and obtain the final model aligned with the target objective. We detail the \algname iterative preference optimization procedure in~\algoref{algo:iterative_preference_optimization}.

\begin{algorithm}[t]
\caption{\algname Iterative Preference Optimization}
\begin{algorithmic}[1]
\REQUIRE Base VLA policy $\pi_\theta$, a collection of task instructions $Q = \{q_i\}$, stage decomposer $\mathcal{M}_D$, max iterations $K$, reward weights $\{\lambda_1, \lambda_2, \lambda_3\}$, stage-wise keypoints $\{\kappa_{S_i}\}$ cost functions $\{C^{S_i}_j\}$ and thresholds $\{\tau^{S_i}_j\}$
\ENSURE policy $\pi^*$ aligned towards customized objective
\FOR{$k = 1, \dots, K$} 
    \STATE Sample trajectories $\mathcal{D}^k = \{\zeta_i\}_{i=1}^M$ using $\pi_\theta$ with $Q$
    \FOR{trajectory $\zeta \in \mathcal{D}^k$}  
        \STATE Decompose $\zeta$ into multiple stages $S$  {\COMMENT{\color{blue} \eqnref{eqn:decomposition}}}
        \STATE Compute the cost for each stage $C_{S_i}$ 
        \STATE Calculate external reward $R_{\text{ext}}(\zeta)$ {\COMMENT{\color{blue} \eqnref{eqn:overall_cost}}}
        \STATE Compute policy self-reward $R_{\text{self}}(\zeta)$  {\COMMENT{\color{blue} \eqnref{eqn:self_traj}}}
        \STATE Examine task success $I_{\text{success}}(\zeta)$ {\COMMENT{\color{blue} \eqnref{eqn:indicator}}}
        \STATE Aggregate GCPG reward $R_{\text{GCPG}}(\zeta)$ {\COMMENT{\color{blue} \eqnref{eqn:overall}}}
    \ENDFOR
    \STATE Rank $\mathcal{D}^k$ by their $R_{\text{GCPG}}(\zeta)$ rewards
    \STATE Pair $\{\zeta_w, \zeta_l\}$ from top-$m$ and bottom-$m$ trajectories
    \STATE Update $\pi_\theta$ using TPO loss {\COMMENT{\color{blue} \eqnref{eqn:tpo}}}
\ENDFOR
\end{algorithmic}
\label{algo:iterative_preference_optimization}
\end{algorithm}

\section{Experiment}

In this section, we evaluate \algname's performance in both real and simulated environments, addressing four key questions: (1) Does \algname improve the VLA model's performance relative to SFT-based baseline models? (2) How effective are guided-cost preference selection and iterative preference optimization in enhancing the model's performance? (3) What is the individual contribution of each reward component to overall model performance? (4) Can \algname support flexible alignment with different alignment objectives?

\subsection{Experimental Setups}
\textbf{Implementation Details.} 
We employ OpenVLA~\cite{kim2024openvla} as the backbone model, using LoRA fine-tuning with the AdamW optimizer for both supervised and preference fine-tuning. In the supervised fine-tuning stage, we use a learning rate of $4 \times 10^{-5}$ with a batch size of 16. For preference fine-tuning, we apply a learning rate of $2 \times 10^{-5}$ with the same batch size. Further details on the training process and datasets are available in Appendices~\ref{sec:app_description} and ~\ref{sec:app_datasets}.

\noindent \textbf{Baseline Models.}
We first compare \algname with two leading robot learning models known for their strong performance in robot control tasks. The first model, Octo~\cite{team2024octo}, is a large transformer-based policy model. The second, OpenVLA~\cite{kim2024openvla}, is a 7B VLA model. Both models were supervised fine-tuned using the same dataset sampled from corresponding environments. We denote the supervised fine-tuned models as Octo-SFT and OpenVLA-SFT, respectively. In addition, we compare \algname, which utilizes trajectory-wise preference optimization, with the original step-wise direct preference optimization, denoted as OpenVLA-DPO, which is directly trained to optimize preferences defined at each step.

\subsection{Evaluation in Simulation Environment}

\noindent \textbf{Evaluation Setup.}
Follow~\citet{kim2024openvla}, we evaluate \algname's performance in two robot simulation environments: Simpler-Env~\cite{li24simpler} and LIBERO~\cite{liu2023libero}. In Simpler-Env, we evaluate the model's in-domain performance as well as its generalization across three aspects: subject (generalize to unseen objects), physical (generalize to unseen object sizes/shapes), and semantic (generalize to unseen instructions) generalization. In LIBERO, we test our model on four tasks: LIBERO-Spatial, LIBERO-Object, LIBERO-Goal, and LIBERO-Long. All tasks are in-domain tasks. Additional details about the experimental setup are provided in Appendix~\ref{sec:detailed_sim_exp}.

\begin{figure}[t!]
    \centering
    \includegraphics[width=0.48\textwidth]{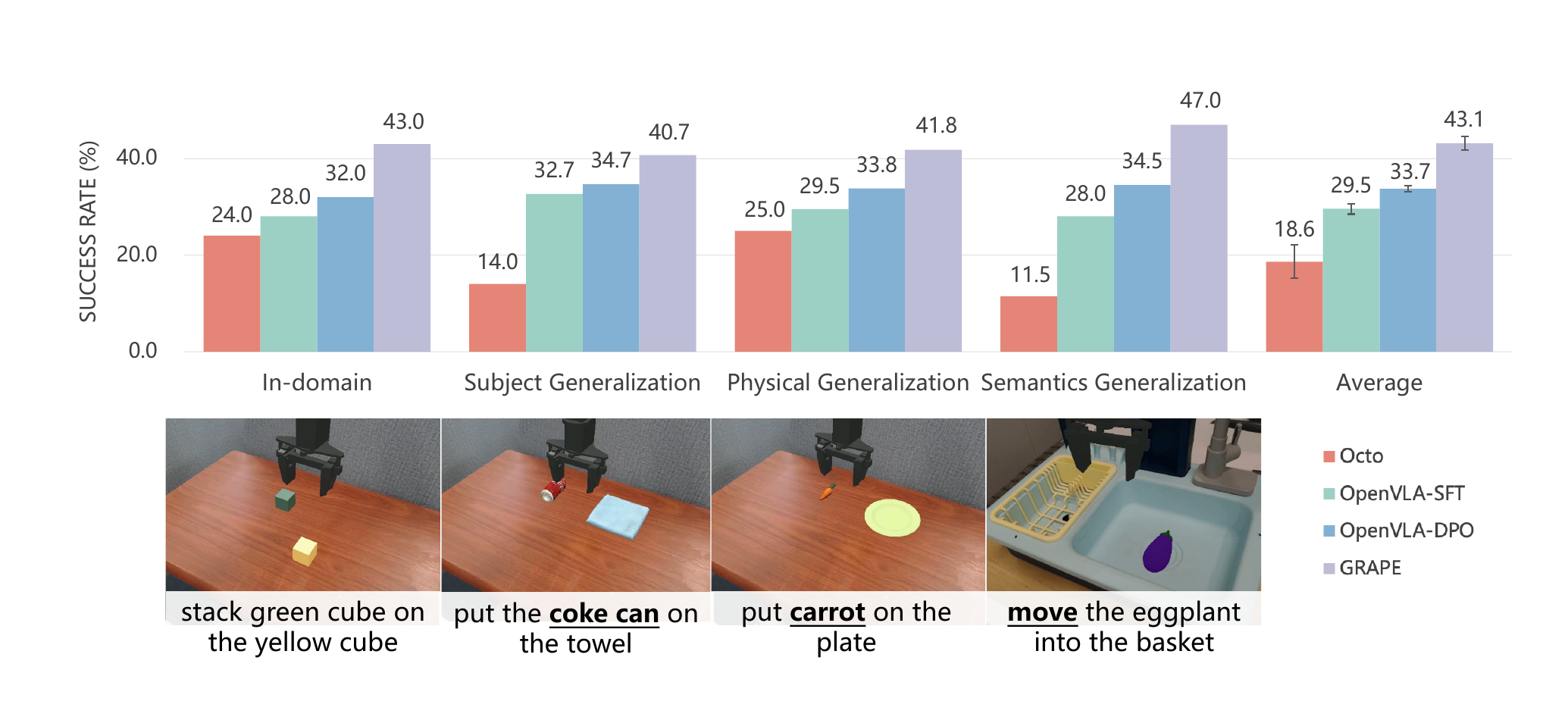}
    \caption{
    Comparison of \algname with OpenVLA and Octo fine-tuned on the same data on the Simpler-Env environment. We report the in-domain performance, which includes four tasks and three generalization evaluations (\textit{subject}, \textit{physical}, and \textit{semantic}), where each incorporates multiple tasks.
}
\label{fig:exp_sim}
\vspace{-1.5em}
\end{figure}

\begin{figure}[t!]
    \centering
    \includegraphics[width=0.48\textwidth]{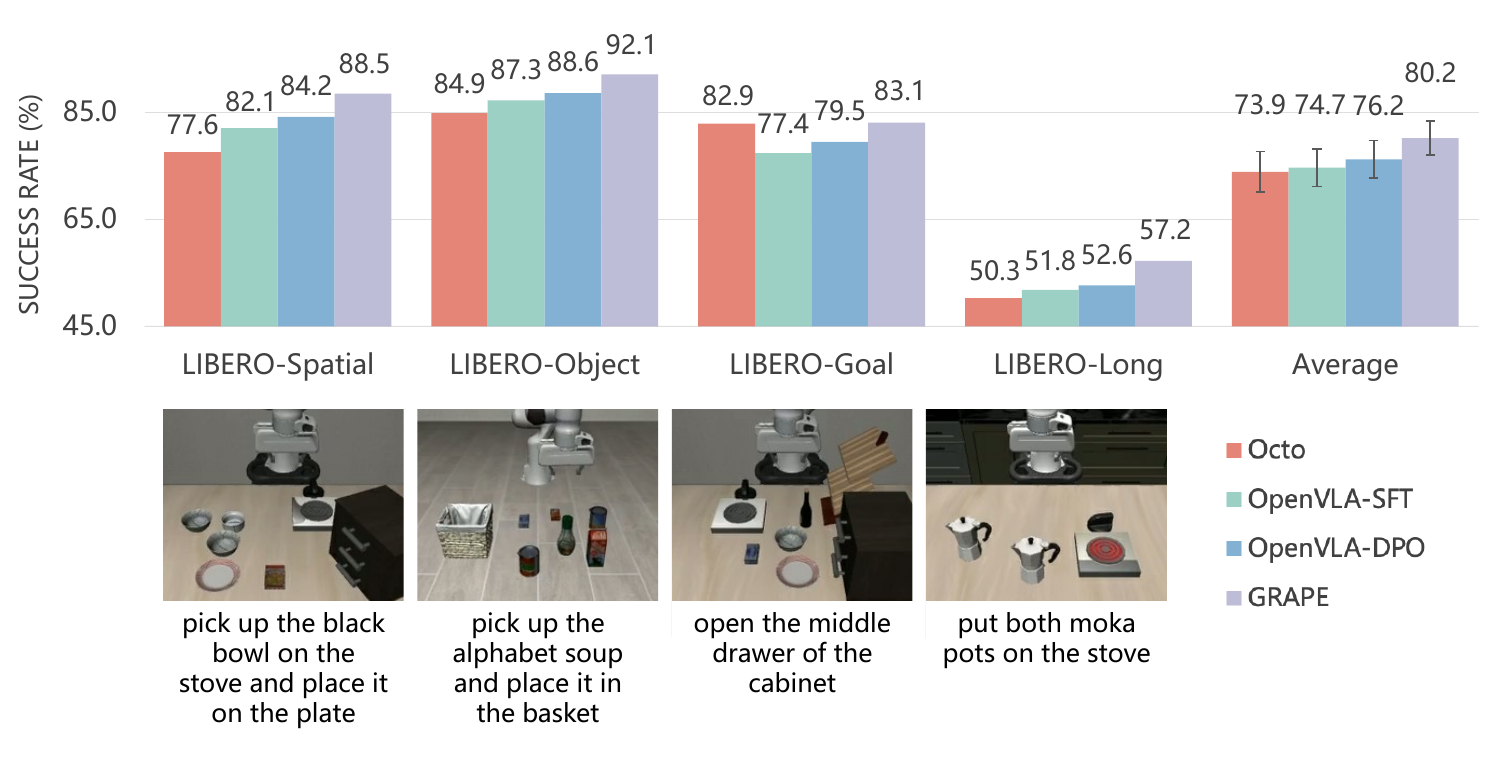}
    \caption{Comparison of \algname with OpenVLA and Octo fine-tuned on the same data on the LIBERO environment. We report the performance on four types of LIBERO tasks.
}
\label{fig:exp_libero}
\vspace{-1.5em}
\end{figure}

\noindent \textbf{Results.}
We use the success rate across all tasks in Simple-Env and LIBERO as our primary evaluation metric, while we also record the grasping rate in Simpler-Env. The results of Simple-Env and LIBERO are reported in~\figref{fig:exp_sim} and~\figref{fig:exp_libero}, respectively. According to the results, \algname outperforms Octo-SFT and OpenVLA-SFT in Simpler-Env by an average of 131.72\% and 46.10\%, respectively, and in LIBERO by an average of 8.53\% and 7.36\%, respectively. Additional results are provided in Appendix~\ref{sec:app_results}. This outcome aligns with our expectations, as learning from preference comparisons enhances alignment with trajectory completion, thereby improving performance. Moreover, while \algname significantly boosts in-domain performance, it also enhances the generalizability of VLA policies on OOD tasks by aligning task completion at the trajectory level. Furthermore, \algname outperforms OpenVLA-DPO in both environments, achieving an average improvement of 33.14\%, demonstrating the effectiveness of trajectory-wise preference optimization due to learning from both success and failure from a global trajectory level without low-level step-wise noises.

\subsection{Evaluation in Real-World Robot Environment}
\begin{figure*}[ht!]
    \centering    \includegraphics[width=1.0\textwidth]{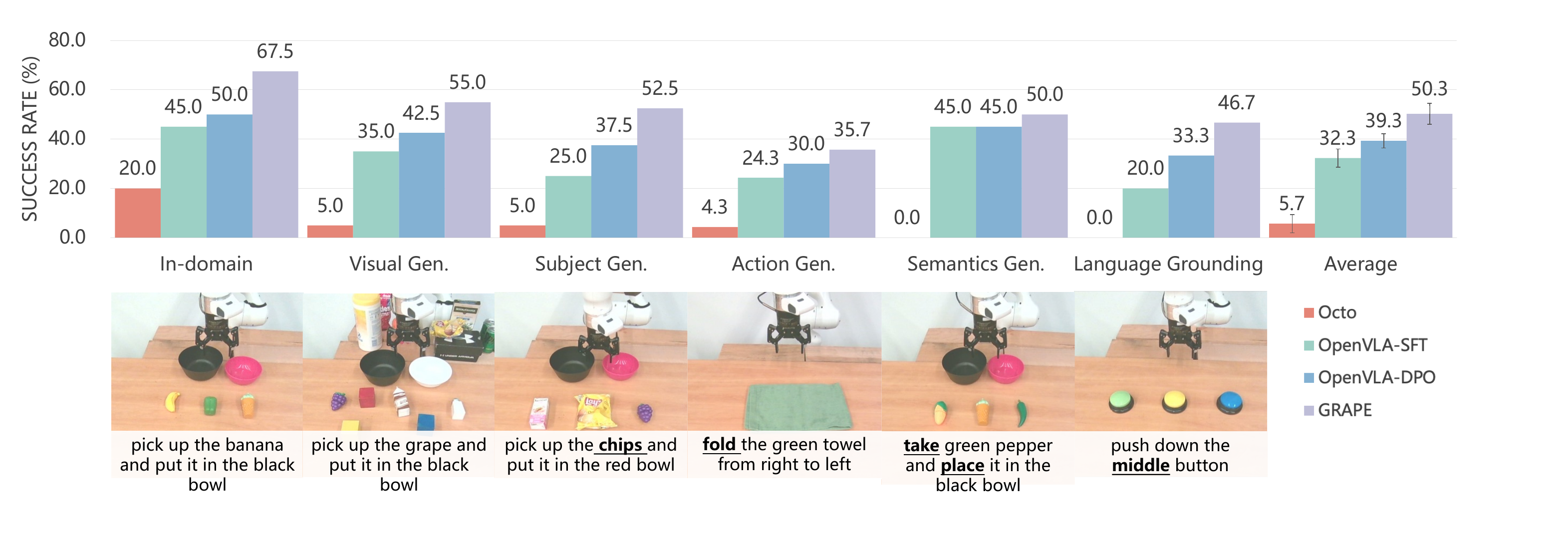}
    \caption{
    Comparison of \algname with OpenVLA and Octo fine-tuned on the same data on the real-world environment. We report the in-domain performance, which includes four tasks and five generalization evaluations (\textit{visual}, \textit{subject}, \textit{action}, \textit{semantic}, and \textit{language grounding}), where each incorporates multiple tasks. We also report the average performance across all tasks.
}
\label{fig:exp_real}
\end{figure*}

\begin{figure*}[h]
    \centering
    \includegraphics[width=0.95\textwidth]{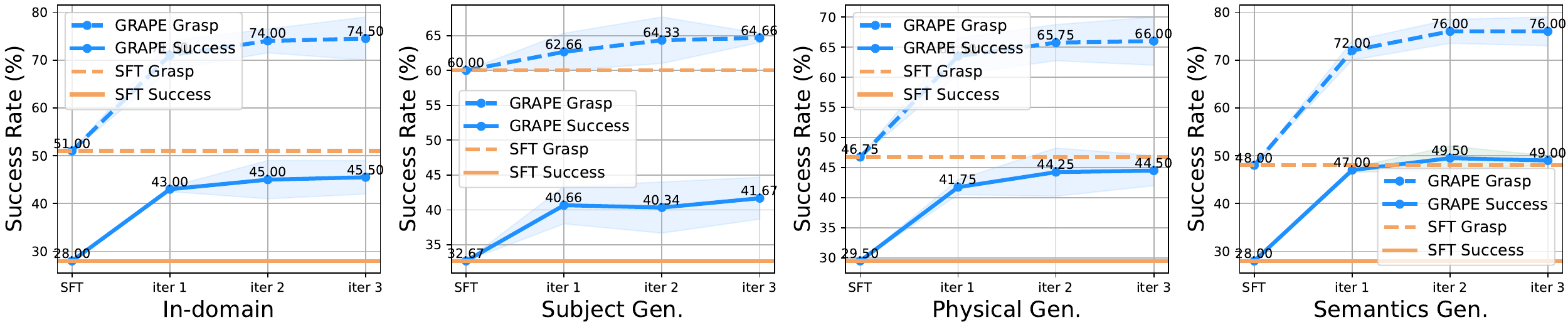}
    \caption{
    Performance of \algname during iterative preference optimization via TPO. We demonstrate the average success rate for each iteration across in-domain tasks and three types of generation tasks (\textit{subject}, \textit{physical}, \textit{semantics}).
}
\label{fig:iteration}
\vspace{-1em}
\end{figure*}
\textbf{Evaluation Setup.} 
We conducted \revision{300} real-world experiments across \revision{30} tasks to evaluate the generalization capabilities of \algname. The evaluation focus on in-distribution evaluation and five out-of-distribution generalization types: visual, subject, action, semantic, and language grounding generalizations. Here, visual generalization assesses the ability to adapt to new visual environments; subject generalization evaluates the recognition and handling of unfamiliar objects; action generalization measures performance across diverse actions; semantic generalization evaluates responses to prompts with similar meanings; and language grounding generalization gauges comprehension of spatial directions. Detailed experimental setup are provided in Appendix~\ref{sec:detailed_realworld_exp} and illustrated in Figure \ref{fig:exp_real}.


\noindent \textbf{Results.}  
In the real-world experiment, \algname significantly outperforms other models across a variety of tasks. Notably, in in-domain tasks, GRAPE achieves a success rate of 67.5\%, which is a \revision{17.5\%} improvement over OpenVLA-DPO's \revision{50\%}, OpenVLA-SFT's \revision{45\%} and substantially higher than Octo-SFT's \revision{20\%}. Additionally, in visual generalization tasks, \algname demonstrates higher adaptability with a success rate of 56\%. In the more challenging action generalization tasks, although OpenVLA-SFT shows modest performance, GRAPE still outperforms OpenVLA-SFT, indicating its potential in understanding various actions and executing commands based on language. Considering tasks across all categories, \algname's total average success rate is \revision{50.3\%}, marking a \revision{11\%} improvement over OpenVLA-DPO's \revision{39.3\%}, OpenVLA-SFT's \revision{32.3\%} and significantly ahead of Octo-SFT's \revision{5.7\%}. This performance highlights (1) \algname's effectiveness and adaptability in handling complex and variable task environments and (2) validates the effectiveness of trajectory-wise preference optimization in learning from global success and failure patterns when compared to OpenVLA-DPO.

\subsection{Ablation Study of Reward Model}
In this section, we conduct an ablation study to analyze the contribution of each reward component in \eqnref{eqn:overall} to the final performance: the external objective-aligned reward \(R_\text{ext}(\zeta)\), the self-evaluated reward \(R_\text{self}(\zeta)\), and the success indicator \(I_{\text{success}}(\zeta)\). Additionally, we perform a separate ablation study to emphasize the importance of utilizing the entire reward score for preference selection. This approach is compared against a method that randomly selects one successful trajectory as the preferred trajectory and one failed trajectory as the rejected trajectory. The results in the Simpler-Env environment are reported in Table \ref{tab:ablation}. 

The results indicate that: (1) incorporating the full reward score \eqnref{eqn:overall} for preference ranking significantly enhances performance compared to random selection based on success alone; (2) all reward components contribute to model performance. These findings align with our expectations. Specifically, \(R_\text{self}(\zeta)\) enhances the robustness of the \algname\ by encouraging it to select trajectories with higher generation probabilities. In parallel, \(R_\text{ext}(\zeta)\) guides the model toward learning specific behaviors, such as safety and efficiency. Finally, \(I_{\text{success}}(\zeta)\) serves as a critical indicator, steering the model to prioritize successful trajectories.

\begin{table*}[h]
\centering
\caption{Ablation study of
reward score. Here, Random w/ $I_\text{success}$ refers to randomly selecting one successful trajectory as the chosen trajectory and one failed trajectory as the rejected trajectory, $R_\text{self}(\zeta)$ is the self-evaluated score provided by the log-likelihood of generating trajectory $\zeta$, $R_\text{ext}(\zeta)$ represents objective-aligned
multi-stage reward defined in \eqnref{eqn:overall_cost}, $I_{\text{success}}(\zeta)$ is a binary indicator function that indicates whether the trajectory $\zeta$ successfully completes the task.}
\label{tab:ablation}
\resizebox{\textwidth}{!}{
\begin{tabular}{l|cc cc cc cc cc}
\toprule
& \multicolumn{2}{c}{\textbf{In-domain}} & \multicolumn{2}{c}{\textbf{Subject Gen.}} & \multicolumn{2}{c}{\textbf{Physical Gen.}} & \multicolumn{2}{c}{\textbf{Semantics Gen.}} & \multicolumn{2}{c}{\textbf{Average}}\\
& Grasp & Success & Grasp & Success & Grasp & Success & Grasp & Success & Grasp & Success\\

\midrule
Random w/ $I_\text{success}$  & 62.00\% & 35.50\% & 60.33\% & 33.00\% & 44.00\% & 33.50\% & 54.50\% & 36.50\% & 55.21\% & 34.63\%\\ 
w/o $R_\text{self}(\zeta)$ & 66.50\% & 38.00\% & 62.33\% & 37.00\% & 51.25\% & 36.75\% & 68.00\% & 42.50\% & 62.02\% & 38.56\%\\  
w/o $R_\text{ext}(\zeta)$ & 63.50\% & 37.50\% & 61.00\% & 34.33\% & 48.50\% & 35.50\% & 62.50\% & 40.00\% & 58.88\% & 36.83\%\\ 
w/o $I_\text{success}$ & 58.50\% & 32.00\% & 59.67\% & 34.67\% & 42.25\% & 31.75\% & 58.50\% & 39.00\% & 54.73\% & 34.36\%\\ 
\midrule
\algname &  \textbf{71.00\%} & \textbf{43.00\%} & \textbf{62.67\%} & \textbf{40.67\%} & \textbf{63.50\%} & \textbf{41.75\%} & \textbf{72.00\%} & \textbf{47.00\%} & \textbf{67.29\%} & \textbf{43.11\%}\\

\bottomrule
\end{tabular}}
\vspace{-1em}
\end{table*}


\subsection{Analysis of Iterative Preference Optimization}
In this section, we analyze the iterative preference optimization performance. We conduct the experiments on the Simpler-Env environment and report the results with respect to the training iterations in Figure~\ref{fig:iteration}. Here, SFT means the supervised fine-tuned OpenVLA model before preference optimization. In our experiments, \algname achieves 17.5\%, 9.0\%, 15.0\%, 21.0\% improvements in in-domain performance, subject generalization, physical generalization and semantic generation, respectively. The findings suggest that \algname progressively enhances model performance across iterations, showcasing its ability to enhance the quality of generated preference data and achieve better generalization. Notably, the magnitude of improvement diminishes over time, aligning with our expectations as the model approaches convergence.



\begin{table}[t]
\footnotesize
    \centering
    \caption{Results with respect to different objectives. \algname-Safety, \algname-Efficiency, \algname-TC are models trained with safety,  efficiency, task completion objectives, respectively. Here, we use collision rate (CR), step length (SL), success rate (SR) to evaluate the safety, efficiency and task completion capabilities.}
    \renewcommand{\arraystretch}{1.2}
    \setlength{\tabcolsep}{3pt} 
    \resizebox{0.45\textwidth}{!}{
    \begin{tabular}{l|ccc|ccc}
        \toprule
        \multirow{2}{*}{Method} & \multicolumn{3}{c|}{\bf Real-World} & \multicolumn{3}{c}{\bf Simulation} \\
        \cmidrule(lr){2-4} \cmidrule(lr){5-7}
        
        & CR $\downarrow$ & SL $\downarrow$ & SR $\uparrow$ & CR $\downarrow$ & SL $\downarrow$ & SR $\uparrow$ \\
        \midrule
        OpenVLA-SFT & 53.33 & 142.32 & 34.61 & 66.50 & 72.68 & 27.50 \\
        \algname-Safety & \revision{\textbf{29.84}} & \revision{146.11} & \revision{54.31} & \textbf{46.00} & 74.49 & 37.00 \\
        \algname-Efficiency & 58.45 & \textbf{125.79} & 51.67 & 57.50 & \textbf{64.92} & 38.50 \\
        \algname-TC & 38.60 & 131.66 & \textbf{58.46} & 59.50 & 70.24 & \textbf{42.50} \\

        \bottomrule
    \end{tabular}
    }
    \label{tab:safety_efficiency_comparison}
    \vspace{-1em}
\end{table}

\subsection{Analysis of Different Alignment Objectives}

\subsubsection{Quantitative Analysis}
After demonstrating the effectiveness of \algname in improving the generalization of the VLA model (measured by success rate), we further investigate its potential to align the model with flexible objectives, such as efficiency and safety. Revisiting \eqnref{eqn:overall_cost}, we observe that adjusting the threshold parameters can guide the model to prioritize specific objectives by influencing trajectory preference selection. In this study, we focus on two new alignment objectives: safety and efficiency. Safety aims to minimize collisions between the robot and objects, while efficiency seeks to reduce the average number of steps required for the robot to complete a task. To achieve these objectives, we set a lower threshold for collision costs to emphasize safety and a lower threshold for path costs to prioritize efficiency. These modified settings are then applied to the original real-world and simulation evaluations. We train models to align with the safety and efficiency objectives, referring to these models as \algname-Safety and \algname-Efficiency, respectively (see detailed experimental setup in Appendix~\ref{sec:detailed_sim_exp}).

The results are reported in~\tabref{tab:safety_efficiency_comparison}, where we use collision rates, step lengths, and success rates to evaluate safety, efficiency and generalization capabilities, respectively. According to~\tabref{tab:safety_efficiency_comparison}, the 
\algname-Safety and \algname-Efficiency have better performance on collision rate and step length respectively, meanwhile maintain a comparable success rate, compared with OpenVLA-SFT. The results indicate that \algname can be easily adapted to account for flexible alignment objectives such as safety, efficiency by adjusting the multi-stage cost functions accordingly, while incurring minimal drop in task success rate.





\subsubsection{Case Study}
\begin{figure*}[ht!]
    \centering
    \includegraphics[width=0.95\textwidth]{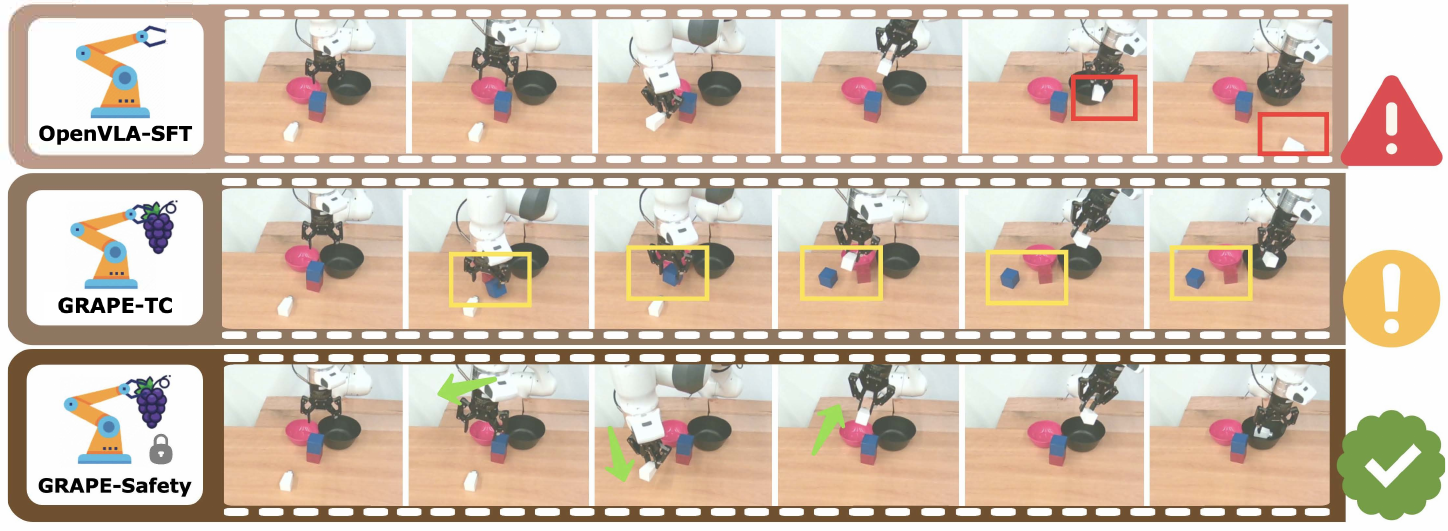}
    \caption{
    Comparison of \algname aligned via \textit{safety} objective (\algname-Safety) with \algname aligned via \textit{task-completion} (\algname-TC) objective and OpenVLA-SFT. Specifically, we assess their performance on a safety-critical task with the instruction: \textit{pick up the white box and place into the black pot}.
}
\label{fig:case_study}
\end{figure*}

We further demonstrate a case study in~\figref{fig:case_study} to analyze \algname's adaptability towards different alignment objectives. Specifically, we consider a safety-critical \textit{pickup} task where an obstacle is placed between the object and the target. Specifically, OpenVLA-SFT fails to complete the task without preference alignment. However, we can see that while \algname aligned towards task completion (on the second-row of~\figref{fig:case_study}) can effectively pick up and place the object, it also collides with the obstacle, due to the policy is aligned to aggressively boost task success without explicitly addressing safety concerns. On the contrary, \algname-safety learns to avoid colliding with the obstacle while efficiently completing the task. Both~\tabref{tab:safety_efficiency_comparison} and~\figref{fig:case_study} indicates that by simply tweaking the cost function, \algname can effectively adapt to different objectives. More cases and detailed safety evaluation tasks could be found in Appendix~\ref{sec:case_study_tasks}.

\section{Related Works}
\label{sec:background}
\vspace{-0.05in}
\textbf{Vision-Language-Action Models.}
Previous robot learning works~\cite{huang2024rekep, li2024hamster, saferlchen, robocodex, huang2023voxposer,liang2023code,mu2024embodiedgpt} typically take a hierarchical planning strategy. For example, Code as Policies~\cite{liang2023code} and EmbodiedGPT~\cite{mu2024embodiedgpt} use LLMs and VLMs to generate high-level action plans, then rely on a low-level controller for local trajectories. However, such models suffer from limited low-level skills and are hard to generalize to everyday tasks.
VLAs tend to scale up low-level tasks by incorporating VLM as backbones and directly generating actions within the model. They generally achieve action planning via two mainstream approaches: 
(1) Discretizing the action space~\cite{kim2024openvla, brohan2023rt, brohan2022rt}, as in OpenVLA~\cite{kim2024openvla}, preserves the autoregressive language decoding objective by truncating actions into a small set of \textit{action tokens}. However, this introduces errors, leading some methods~\cite{black2024pi_0} to adopt newer structures~\cite{zhou2024transfusion} that integrate diffusion heads for action prediction, avoiding discretization. (2) Diffusion models~\cite{chi2023diffusion, xian2023chaineddiffuser, diffuser, adaptdiffuser, decisiondiffuser}, such as Diffusion Policy~\cite{chi2023diffusion}, serve as the action head, generating a sequence of future actions through iterative denoising instead of stepwise action generation.

While these models vary in structure, they are consistently supervised-trained on successful rollouts via behavior cloning, which can hardly be generalized to unseen manipulation tasks. However, our \algname first aligns VLA policies on a trajectory level via trial and error, effectively boosting generalizability and customizability.

\vspace{-0.05in}
\noindent \textbf{Reinforcement Learning and Preference Optimization.} Reinforcement learning (RL)~\cite{christiano2017deep,ziegler2019fine,schulman2017proximal} plays a pivotal role in the post-training of foundation models~\cite{dubey2024llama, achiam2023gpt, chen2024autoprm, chen2024mj, fan2024reinforcement, yang2024cogvideox, chen2024safewatch, wang2025beyond}, which has been extensively leveraged to align the pre-trained FMs to comply with human values embedded through preference data. 
In the meantime, RL has also shown tremendous success in training policies for robotics tasks~\cite{saferlchen, wang2024escirl, chen2021decision,chen2022towards,RLdexterousReview,wu2024unidexfpm}. While it is intuitively beneficial to post-align VLA via RL, few prior works have reported such success, mainly due to that (1) manipulation objectives are usually diverse and complex, making the reward hard to define analytically~\cite{finn2016guided}; (2) while such reward can be modeled from human preferences, annotating such preferences in robotics manipulation tasks are usually lengthy~\cite{walke2023bridgedata}; (3) the imperfect numerical differentiation of rewards usually leads RL algorithms such as PPO~\cite{schulman2017proximal} to collapse~\cite{bucsoniu2018reinforcement}.
However, various recent works~\cite{rafailov2024direct,wang2024preference} have successfully aligned the policy via RL without explicit reward modeling. Inspired, \algname aligns the policy by contrasting trajectories with each other, avoiding issues in rewarding modeling. Besides, we introduce an automatic preference synthesis pipeline that easily scales with diverse manipulation tasks and adapts to different alignment objectives.

\vspace{-0.5em}
\section{Conclusion}
\vspace{-0.5em}
In this work, we addressed the critical challenges faced by vision-language-action (VLA) models, including limited generalizability and adaptability to diverse manipulation objectives. We proposed \algname, which aligns VLA policies on a trajectory level. \algname enhances generalizability by learning from both successful and failed trials, offering flexibility in aligning with objectives such as safety, efficiency, and task success through customized spatiotemporal constraints. Experimental results demonstrated significant improvements, with GRAPE enhancing success rates on both in-domain and unseen tasks while enabling flexible alignment on different objectives. Moreover, we have demonstrated the potential of GRAPE to align VLA with customized objectives, effectively resulting in an improvement of lower collision rate and average step lengths.

\section*{Impact Statement}
This paper presents work whose goal is to advance the field of 
Machine Learning. There are many potential societal consequences 
of our work, none which we feel must be specifically highlighted here.

\nocite{langley00}

\bibliography{main}
\bibliographystyle{icml2025}

\newpage
\appendix
\onecolumn

\clearpage

\appendix
\onecolumn
\section{Additional Description of \algname and Hyperparameter Settings}
\label{sec:app_description}

\textbf{Customized Cost Generation.}
In our real-world experiments, we first input image-text pairs containing prompts and initial states into the Vision-Language Model (VLM) Hamster~\cite{li2024hamster}. Using the stage information and stage points generated by Hamster, we segmented the collected trajectories. This helps analyze complex task sequences more precisely, giving detailed attention to each stage. And we utilized Grounded-SAM~\cite{ravi2024sam2} or methods combining SAM~\cite{ravi2024sam2} and DinoV2~\cite{oquab2024dinov2learningrobustvisual} to extract key point information from the images. These key points, combined with our self-collected trajectory data, enable us to refine the execution steps and path planning of tasks based on the stage information generated by the Hamster model. For example, for a simple pick-and-place task, we can decompose it into multiple explicit stages: Grasp the grape, Move the grape onto the plate, Place the grape on the plate.

To generate detailed operational information and cost functions for each stage, we utilized GPT-4o~\cite{achiam2023gpt} with customized prompts. This approach makes stage planning more precise and efficient, allowing us to meet specific task requirements and constraints. Furthermore, we enhanced our method by incorporating various task-specific constraints, including: \textbf{Collision constraints:} Ensuring the robot avoids collisions with obstacles. \textbf{Path constraints:} Optimizing the efficiency and safety of the robot's movement path. By adopting this strategy, we achieve greater flexibility and specificity in task planning, and better adapting to different task scenarios.


\noindent \textbf{Iterative Preference Optimization.}
For Iterative Preference Optimization, we first utilize the fine-tuned VLA model for online data sampling. For each task, we sample $\mathcal{N}_t$ trajectories to facilitate further selection. To simplify the experimental setup, we set $\mathcal{N}_t = 5$ for each task, which has been found to perform effectively in practice.

After sampling, each trajectory is automatically labeled using the GCPG reward, as defined in Eq.~\eqref{eqn:overall}. Based on the distribution of $R_\text{self}$, $R_\text{ext}$, and $I_\text{success}$ observed in preliminary experiments, we set $\lambda_1 = 0.01$, $\lambda_2 = 0.01$, and $\lambda_3 = 2$. These values ensure that $R_\text{self}$, $R_\text{ext}$, and $I_\text{success}$ contribute comparably to the final reward value. Subsequent experiments validate the reasonableness of these parameter choices. Using the GCPG reward assigned to each trajectory, we identify the trajectory with the highest reward as $y_{w}$ and the trajectory with the lowest reward as $y_{l}$ for each task. This selection process enables the construction of the TPO Dataset, $\mathcal{D}_{traj}$, for TPO training.

For the TPO training process, we employ LoRA~\cite{hu2022lora} and the AdamW optimizer, setting the learning rate to $2 \times 10^{-5}$ and the batch size to 16. The model is trained for a single epoch before being utilized for iterative online sampling. During iterative online sampling, the experimental settings remain consistent with the aforementioned descriptions.

\section{Detail Experiment Datasets}
\label{sec:app_datasets}
In this section, we describe the datasets collected for supervised fine-tuning (referred to as the SFT dataset) and preference alignment (referred to as the TPO dataset).

\subsection{Real-World Dataset}

\textbf{SFT Dataset.}
In our real-world robot experiments, we use a robotic platform composed of a Franka robotic arm and a Robotiq gripper for data collection. To ensure consistency in data collection and evaluation, all operations are performed in the same experimental environment.

During data collection, we gathered a dataset of 220 instances of pick and place tasks involving common objects such as bananas, corn, milk, and salt. Additionally, we collected data on 50 instances of tasks involving pressing buttons of different colors. Since the number of objects used for the button-pressing tasks is limited, we introduced background noise and interfering objects during the testing phase to create unseen scenarios.

To further enhance the capabilities of OpenVLA in handling different actions, we also collected data on 50 instances of knock down tasks. These diverse task datasets help improve the model's generalization ability in processing different types of actions.

\noindent \textbf{TPO Dataset.}
In the real-world experiments, we utilized a model fine-tuned on the real-world SFT dataset via OpenVLA for trajectory sampling. Each task was conducted five times. In the TPO dataset, we experimented with 15 different tasks, including 10 pick and place tasks, 3 push button tasks, and 2 knock down tasks, accumulating a total of 75 data entries. After a selection process, we derived a preference dataset consisting of 30 trajectories.

\subsection{Simulation Datasets}
\noindent \textbf{SFT Dataset:} For Simpler-Env, the SFT dataset comprises 100 trajectories, amounting to approximately 2,900 transitions. These rollouts are generated from Simpler-Env using Octo, following the methodology described in \citet{ye2024latent}. For LIBERO, it is worth noting that we neither collect new data nor fine-tune the OpenVLA model. Instead, we directly utilize the OpenVLA-SFT model provided by the OpenVLA team, which significantly streamlines the pipeline.

\noindent \textbf{TPO Dataset.} In the case of Simpler-Env, trajectories are sampled for each task using the OpenVLA-SFT model, with five trials conducted per task. This process yields a TPO dataset consisting of 80 trajectories. For LIBERO, OpenVLA-SFT models (one model per task) are employed to sample data across four tasks in LIBERO. For each task, five trajectories are sampled for each sub-task, resulting in a TPO dataset comprising a total of 20 trajectories.

\section{Detailed Experiment Settings and Additional Result}
\label{sec:app_experiment}
\subsection{Real-World}
\label{sec:detailed_realworld_exp}
\subsubsection{Real-World Experiment Setup}

In real-world experiment, we used the Franka robot arm, which is known for its precision and flexibility. However, we encountered a problem with the original Franka gripper, which was not long enough, limiting our ability to handle some of the tasks, resulting in inefficient completion and a high failure rate. To solve this problem, we decided to replace the original Franka grippers with Robotiq grippers, which are not only longer, but also provide more grip and flexibility, which greatly improves the efficiency and success rate of the tasks.

The purpose of this experiment was to assess the cross-task generalization capabilities of OpenVLA under the \algname framework and to compare its performance with several baseline models. Considering the generally poor zero-shot generalization performance of most VLA models, we performed supervised fine-tuning using the comprehensive rollout dataset $D_r$ collected from real scenes to construct a fine-tuned model. The selection of baseline models included those adjusted with domain-specific data, as well as the Octo model, RVT-2 model, and OpenVLA-SFT model.

\subsubsection{Real-World Tasks}
As shown in Figure \ref{fig:exp_real}, we performed a comprehensive evaluation on a real machine for several tasks. These tasks cover five different generalization scenarios: Visual Generalization, Subject Generalization, Action Generalization, Semantics Generalization, and Language Grounding. Specifically, for each generalization scenario, we set the following tasks:

\begin{itemize}
    \item \textbf{Visual Generalization} includes 8 tasks, e.g., pick up the \algname and put it in the black bowl, with noise objects and noisy backgrounds. Some tasks have only noisy backgrounds.
    \item \textbf{Subject Generalization} includes 4 tasks, e.g., pick up the K and put it in the black bowl.
    \item \textbf{Action Generalization} includes 7 tasks, e.g., fold the green towel from right to left .
    \item \textbf{Semantics Generalization} includes 4 tasks, e.g., stack carrot and put it on the blue plates.
    \item \textbf{Language Grounding} includes 3 tasks, e.g., pick up left object to left plate.
\end{itemize}

We conducted experiments on 30 total different tasks, attempting each task ten times, totaling 300 executions. To ensure fairness in the evaluation, we maintained the same starting position in each model test. Additionally, we matched the image resolution when training all models and used exactly the same initial object positions in all evaluations. We set specific success criteria for each task. For example, in the pick-and-place task, a successful grasp is defined as successfully grasping the target object. In the push-button and knock-down tasks, a successful grasp is defined as correctly approaching and manipulating the target object. Overall task success is defined as the object being accurately placed at the target location, successfully knocked down, or the target button being successfully pressed. Due to the strictness of these criteria, some models found it difficult to achieve success in specific tasks.

\begin{table*}[h]
\centering
\caption{Comparison of \algname models in diffierent iteration rounds. We assess their performance in in-domain tasks and three kinds of generalization evaluations. Each task's performance is evaluated on the overall grasp rate and success rate.}
\label{tab:performance_robotics}
\small
\begin{tabular}{l|cc cc cc cc cc}
\toprule
& \multicolumn{2}{c}{\textbf{In-domain}} & \multicolumn{2}{c}{\textbf{Subject Gen.}} & \multicolumn{2}{c}{\textbf{Physical Gen.}} & \multicolumn{2}{c}{\textbf{Semantics Gen.}} & \multicolumn{2}{c}{\textbf{Average}}\\
& Grasp & Success & Grasp & Success & Grasp & Success & Grasp & Success & Grasp & Success\\

\midrule
Iter-1  & 71.00\% & 43.00\% & 62.67\% & \textbf{40.67}\% & 63.50\% & 41.75\% & 72.00\% & 47.00\% & 67.29\% & 43.11\%\\ 
Iter-2 & 74.00\% & 45.00\% & 64.33\% & 40.33\% & 65.75\% & 44.25\% & \textbf{76.00\%} & \textbf{49.50\%} & 70.02\% & 44.77\% \\ 
Iter-3 & \textbf{74.50\%} & \textbf{45.50\%} & \textbf{64.67\%} & \textbf{40.67\%} & \textbf{66.00\%} & \textbf{44.50\%} & \textbf{76.00\%} & 49.00\% & \textbf{70.29\%} & \textbf{44.92\%}\\

\bottomrule
\end{tabular}
\end{table*}

\subsection{Simulation Experiments}
\label{sec:detailed_sim_exp}
\subsubsection{Simpler-Env}
We utilize Simpler-Env~\cite{li24simpler} as the experimental environment in our study. SIMPLER~\cite{li24simpler} (Simulated Manipulation Policy Evaluation for Real Robot Setups) is a collection of simulated environments created to assess robot manipulation policies in a way that closely reflects real-world scenarios. By leveraging simulated environments, SIMPLER effectively serves as a practical alternative to real-world testing, which is often costly, time-consuming, and challenging to replicate.

\noindent \textbf{Simpler-Env Tasks.} In our paper, we use four in-domain tasks from WidowX robot in Simpler-Env. We also design three kinds of generalization tasks in Simpler-Env. These tasks are described below:

\noindent \textbf{In-Domain Tasks Shown in Fig.~\ref{fig:exp_sim}:}
\begin{enumerate}
    \item Put Carrot on Plate: The robot is positioned in front of a platform with a plate and a carrot. The robot’s goal is to grasp the carrot and put it onto the plate.
    \item Put Eggplant in basket: The robot is positioned in front of a sink with a basket and a Eggplant. The robot’s goal is to grasp the Eggplant and put it in the basket.
    \item Stack Green Cube on Yellow Cube: The robot is positioned in front of a platform with a green cube and a yellow cube. The robot’s goal is to grasp the green cube and stack it on the yellow cube.
    \item Put Spoon on towel: The robot is positioned in front of a platform with a spoon and a towel. The robot’s goal is to grasp the spoon and put it on the towel.
\end{enumerate}

\noindent \textbf{Three Kinds of Generalization Tasks Shown in Fig.~\ref{fig:exp_sim}:}
\begin{enumerate}
    \item Subject Generalization: The robot is positioned in front of a platform, similar to the environment in in-domain tasks. But the robot's goal is to grasp some new objects(i.e. pepsi can, coke can, sprite can) and put it onto the plate.
    \item Physical Generalization: The robot is positioned in front of a platform, similar to the environment in in-domain tasks. But the robot's goal is to grasp some original objects with different sizes and collision boxes, then put it onto the plate.
    \item Semantics Generalization: The robot is positioned in front of a platform, similar to the environment in in-domain tasks. And the instruction is similar to in-domain tasks, too. But the instruction has been modified by GPT-4o~\cite{achiam2023gpt} while maintaining its original meaning.
\end{enumerate}

\subsubsection{LIBERO}
We further utilize LIBERO~\cite{liu2023libero} as the experimental environment in our study. LIBERO (LIfelong learning BEnchmark on RObot manipulation tasks) includes a set of 130 language-conditioned robot manipulation tasks inspired by human activities, organized into four distinct suites. Each suite is crafted to examine distribution shifts in object types, spatial arrangements of objects, task goals, or a combination of these factors. LIBERO is built to be scalable, extendable, and specifically tailored for advancing research in lifelong learning for robotic manipulation.

\noindent \textbf{LIBERO tasks} In our paper, we use four in-domain tasks from LIBERO, which are shown in Fig.~\ref{fig:exp_libero}. These tasks is described below:
\begin{itemize}
    \item \textbf{LIBERO-Spatial} includes the same set of objects arranged in various layouts, testing the model's ability to understand spatial relationships.
    \item \textbf{LIBERO-Object} features consistent scene layouts with varying objects, evaluating the model's ability to understand different object types.
    \item \textbf{LIBERO-Goal} includes of the same objects and layouts but different task goals, testing the model’s knowledge of different task-oriented behaviors.
    \item \textbf{LIBERO-10} consists of long-horizon tasks with diverse objects, layouts, and tasks.
\end{itemize}
Eash task mentioned above has 10 sub-tasks, with similar task instructions and scenes. Here are some cases from various LIBERO tasks:
\begin{itemize}
    \item Open the top drawer of the cabinet and put the bowl in it.
    \item Pick up the book and place it to the right of the caddy.
    \item Turn on the stove and put the frying pan on it.
    \item Stack the right bowl on the left bowl and place them in the tray.
\end{itemize}

\section{Additional Real-World and Simulation Results}
\label{sec:app_results}
We provide additional results in Table \ref{tab:real_world_robotics} , Table \ref{tab:Simpler-Env_robotics}, and Figure \ref{fig:real-world-case-6} with detailed task description. Each table has in-domain tasks and several kinds of generalization evaluations. These experiments are conducted across Octo-SFT, OpenVLA-SFT and \algname.

\begin{table*}[h]
\centering
\caption{We present the performance of various action policy on real-world robotic manipulation tasks categorized by different types of generalization. The tasks include in-domain, visual generalization with and without noise, subject generalization, action generalization, semantics generalization, and language grounding. Each task's performance is evaluated based on the number of successful grasps and the overall success rate, comparing results from Octo-SFT, OpenVLA-SFT, \revision{OpenVLA-DPO}, and \algname. Average success rates are calculated for each generalization category to demonstrate the effectiveness of the tested models under different conditions.}
\label{tab:real_world_robotics}
\resizebox{\textwidth}{!}{
\begin{tabular}{l|l| cc cc cc cc}
\toprule

\textbf{Generalization} & \textbf{Task} & \multicolumn{2}{c}{\textbf{Octo-SFT}} & \multicolumn{2}{c}{\textbf{OpenVLA-SFT}} & \multicolumn{2}{c} {\textbf{OpenVLA-DPO}} & \multicolumn{2}{c} {\textbf{\algname}} \\
& & Grasp & Success & Grasp & Success & Grasp & Success & Grasp & Success  \\

\midrule
\multirow{5}{*}{In-domain} & pick up the corn and put it in the black bowl & 3 & 3 & 2 & 2 & 5 & 3 & 8 & 7 \\
& pick up the banana and put it in the black bowl & 2 & 0 & 6 & 6 & 8 & 6 & 9 & 7 \\
& pick up the milk and put it in the white bowl & 4 & 2 & 10 & 8 & 8 & 8 & 9 & 9 \\
& pick up the salt bottle and put it in the white bowl & 4 & 3 & 4 & 2 & 5 & 3 & 6 & 4 \\
& Average & 32.5\% & 20\% & 55\% & 45\% & 65\% & 50\% & \textbf{80\%} & \textbf{67.5\%} \\

\midrule
\multirow{6}{*}{\parbox{5cm}{Visual Generalization \\ (w/o noise background)}}  & pick up the corn and put it in the black bowl & 2 & 1 & 6 & 3 & 6 & 4 & 6 & 6 \\
& pick up the banana and put it in the black bowl & 0 & 0 & 3 & 2 & 4 & 1 & 4 & 1 \\
& pick up the milk and put it in the white bowl & 4 & 0 & 4 & 4 & 6 & 6 & 9 & 7 \\
& pick up the salt bottle and put it in the white bowl & 2 & 2 & 6 & 5 & 6 & 6 & 8 & 8 \\
& pick up the \algname and put it in the black bowl & 0 & 0 & 6 & 5 & 8 & 5 & 8 & 6 \\
& Average & 16\% & 6\% & 50\% & 38\% & 60\% & 44\% & \textbf{70\%} & \textbf{56\%} \\

\midrule
\multirow{4}{*}{\parbox{5cm}{Visual Generalization \\ (w/o noise background and object)}} & pick up the \algname and put it in the black bowl & 1 & 0 & 4 & 2 & 5 & 3 & 6 & 4 \\
& pick up the milk and put it in the white bowl & 2 & 1 & 7 & 5 & 6 & 4 & 5 & 4 \\
& pick up the salt bottle and put it in the white bowl & 0 & 0 & 2 & 2 & 6 & 5 & 8 & 8 \\
& Average & 10\% & 3.3\% & 43.3\% & 30\% &56.7\% & 40\% & \textbf{63.3\%} & \textbf{53.3\%} \\

\midrule
\multirow{5}{*}{Subject Generalization)} & pick up the chips and put it in the red bowl & 4 & 0 & 2 & 2 & 4 & 3 & 6 & 5 \\
& pick up the K and put it in the black bowl & 2 & 0 & 4 & 4 & 6 & 5 & 7 & 6 \\
& pick up the box juice and put it in the yellow plate & 2 & 0 & 8 & 3 & 8 & 5 & 8 & 6 \\
& pick up the Fanta can and put it in the white bowl & 2 & 2 & 4 & 1 & 5 & 2 & 6 & 4 \\
& Average & 25\% & 5\% & 45\% & 25\% & 57.5\% & 37.5\% & \textbf{67.5\%} & \textbf{52.5\%} \\

\midrule
\multirow{8}{*}{Action Generalization} & push down the blue button & 1 & 0 & 4 & 4 & 6 & 4 & 6 & 6 \\
& push down the green button & 1 & 0 & 6 & 4 & 7 & 5 & 4 & 4 \\
& push yellow the button & 2 & 2 & 6 & 3 & 7 & 4 & 8 & 5 \\
& knock down the green bottle & 3 & 1 & 2 & 2 & 3 & 2 & 4 & 2 \\
& knock down the popcorn & 0 & 0 & 4 & 2 & 4 & 3 & 4 & 3 \\
& \revision{fold the green towel from right to left} & 1 & 0 & 2 & 1 & 3 & 1 & 4 & 2 \\
& \revision{fold the white towel from left to right} & 1 & 0 & 3 & 1 & 4 & 2 & 4 & 3 \\
& Average & 12.9\% & 4.3\% & 38.6\% & 24.3\% & \textbf{48.6\%} &30\% & \textbf{48.6\%} & \textbf{35.7\%} \\

\midrule
\multirow{5}{*}{Semantics Generalization} & take green pepper and place it in the black bowl & 0 & 0 & 10 & 6 & 9 & 7 & 10 & 8 \\
& move icecream and put it in the red bowl & 0 & 0 & 6 & 4 & 5 & 4 & 4 & 4 \\
& stack carrot and put it on the blue plates & 0 & 0 & 8 & 8 & 6 & 5 & 6 & 6 \\
& Lift \algname and place it in the black bowl & 0 & 0 & 2 & 0 & 3 & 2 & 2 & 2 \\
& Average & 0\% & 0\% & \textbf{65\%} & 45\% & 57.5\% &45\% & 55\% & \textbf{50\%} \\

\midrule
\multirow{4}{*}{Language Grounding} & pick up left object to left plate & 0 & 0 & 4 & 0 & 5 & 1 & 5 & 2 \\
& push down right button & 0 & 0 & 6 & 2 & 6 & 5 & 8 & 7 \\
& pick up right object to right plate & 0 & 0 & 4 & 4 & 5 & 4 & 6 & 5 \\
& Average & 0\% & 0\% & 46.7\% & 20\% &53.3\% & 33.3\% & \textbf{63.3\%} & \textbf{46.7\%} \\


\midrule
Total Average & & 14.3\% & 5.7\% & 48.3\% & 32.3\% &56.3\% &39.3\% & \textbf{62.6\%} & \textbf{50.3\%}\\

\bottomrule
\end{tabular}}
\end{table*}

\begin{table*}[t!]
\centering
\caption{\revision{We compared the performance of Octo-SFT, OpenVLA-SFT, and \algname across various robotic tasks within in-domain, subject, physical, and semantics generalization categories. It shows grasp percentages and success rates for each task, illustrating how each VLA performs under different generalizations.}}
\label{tab:Simpler-Env_robotics}
\resizebox{\textwidth}{!}{
\begin{tabular}{l|l|cc cc cc cc}
\toprule
\textbf{Generalization} & \textbf{Task} & \multicolumn{2}{c}{\textbf{Octo-SFT}} & \multicolumn{2}{c}{\textbf{OpenVLA-SFT}} & \multicolumn{2}{c}{\textbf{OpenVLA-DPO}} & \multicolumn{2}{c}{\textbf{\algname}} \\
& & Grasp & Success & Grasp & Success & Grasp & Success & Grasp & Success \\

\midrule
\multirow{5}{*}{In-domain} & put the carrot on the plate & 32.00\% & 16.00\% & 36.00\% & 30.00\% & 46.00\% & 36.00\% & \textbf{68.00\%} & \textbf{48.00\%} \\
& put the eggplant in the basket & 70.00\% & 44.00\% & 58.00\% & 32.00\%  & 70.00\% & 36.00\% & \textbf{84.00\%} & \textbf{48.00\%} \\
& stack the green cube on the yellow cube & 52.00\% & 0.00\% & 56.00\% & 20.00\%  & 52.00\% & 26.00\% & \textbf{76.00\%} & \textbf{40.00\%} \\
& put the spoon on the towel & 54.00\% & \textbf{36.00\%} & 52.00\% & 28.00\%  & 52.00\% & 30.00\% & \textbf{56.00\%} & 34.00\% \\
& Average & 52.00\% & 24.00\% & 50.50\% & 28.00\% & 55.00\% & 32.00\% & \textbf{71.00\%} & \textbf{43.00\%} \\

\midrule
\multirow{4}{*}{\parbox{4cm}{Subject Generalization \\ (unseen objects)}} & put the coke can on the towel & 24.00\% & 14.00\% & 60.00\% & \textbf{38.00\%} & 66.00\% & 36.00\% & \textbf{78.00\%} & 32.00\% \\
& put the pepsi can on the towel & 28.00\% & 16.00\% & 58.00\% & 38.00\% & 60.00\% & 42.00\% & \textbf{64.00\%} & \textbf{50.00\%} \\
& put the sprite can on the towel & 24.00\% & 12.00\% & \textbf{62.00\%} & 22.00\% & 58.00\% & 26.00\% & 46.00\% & \textbf{40.00\%} \\
& Average & 25.33\% & 14.00\% & 60.00\% & 32.67\% & 61.33\% & 34.66\% & \textbf{62.67\%} & \textbf{40.67\%} \\

\midrule
\multirow{9}{*}{\parbox{4cm}{Physical Generalization \\ (unseen object sizes/shapes)}} & put the carrot on the plate(size:0.5) & 38.00\% & 22.00\% & 56.00\% & 38.00\% & 60.00\% & 46.00\% & \textbf{78.00\%} & \textbf{64.00\%} \\
& put the carrot on the plate(size:1.1) & 26.00\% & 12.00\% & 32.00\% & 24.00\% & 42.00\% & 30.00\% & \textbf{64.00\%} & \textbf{42.00\%} \\
& put the carrot on the plate(wider collision box) & 28.00\% & 16.00\% & 34.00\% & 26.00\% & 46.00\% & 32.00\%  & \textbf{62.00\%} & \textbf{42.00\%} \\
& put the carrot on the plate(longer collision box) & 32.00\% & 14.00\% & 38.00\% & 30.00\% & 50.00\% & 36.00\% & \textbf{66.00\%} & \textbf{48.00\%} \\
& put the spoon on the towel(size:0.5) & 62.00\% & 38.00\% & 66.00\% & \textbf{40.00\%} & 66.00\% & 38.00\% & \textbf{72.00\%} & 38.00\% \\
& put the spoon on the towel(size:1.1) & 52.00\% & \textbf{32.00\%} & 50.00\% & 28.00\% & \textbf{58.00\%} & \textbf{32.00\%} & 56.00\% & 30.00\% \\
& put the spoon on the towel(wider collision box) & 48.00\% & 30.00\% & 44.00\% & 24.00\% & 46.00\% & 28.00\% & \textbf{50.00\%} & \textbf{32.00\%} \\
& put the spoon on the towel(longer collision box) & 56.00\% & 36.00\% & 54.00\% & 26.00\% & 54.00\% & 28.00\% & \textbf{60.00\%} & \textbf{38.00\%} \\
& Average & 42.75\% & 25.00\% & 46.75\% & 29.50\% & 52.75\% & 33.75\% & \textbf{63.50\%} & \textbf{41.75\%} \\

\midrule
\multirow{4}{*}{\parbox{4cm}{Semantics Generalization \\ (unseen instructions)}} & put the vegetable on the plate & 16.00\% & 6.00\% & 32.00\% & 28.00\% & 40.00\% & 32.00\% & \textbf{66.00\%} & \textbf{48.00\%} \\
& move the eggplant into the basket & 18.00\% & 8.00\% & 50.00\% & 30.00\% & 56.00\% & 34.00\% & \textbf{78.00\%} & \textbf{44.00\%} \\
& put the green cube onto the yellow cube & 32.00\% & 6.00\% & 62.00\% & 26.00\% & 74.00\% & 42.00\% & \textbf{88.00\%} & \textbf{60.00\%} \\
& place the spoon onto the towel & 42.00\% & 26.00\% & 48.00\% & 28.00\% & 48.00\% & 30.00\% & \textbf{56.00\%} & \textbf{36.00\%} \\
& Average & 27.00\% & 11.50\% & 48.00\% & 28.00\% & 54.50\% & 34.50\% & \textbf{72.00\%} & \textbf{47.00\%} \\

\midrule
\multirow{1}{*}{Total average} &  & 36.77\% & 18.63\% & 51.44\% & 29.54\% & 55.90\% & 33.73\% & \textbf{67.29\%} & \textbf{43.11\%} \\

\bottomrule
\end{tabular}}
\end{table*}

\section{Case Study}

\subsection{Case Study of Real-World Generation Tasks}
\label{sec:case_study_tasks}
\begin{figure*}[htbp]
    \centering
    \includegraphics[width=0.95\textwidth]{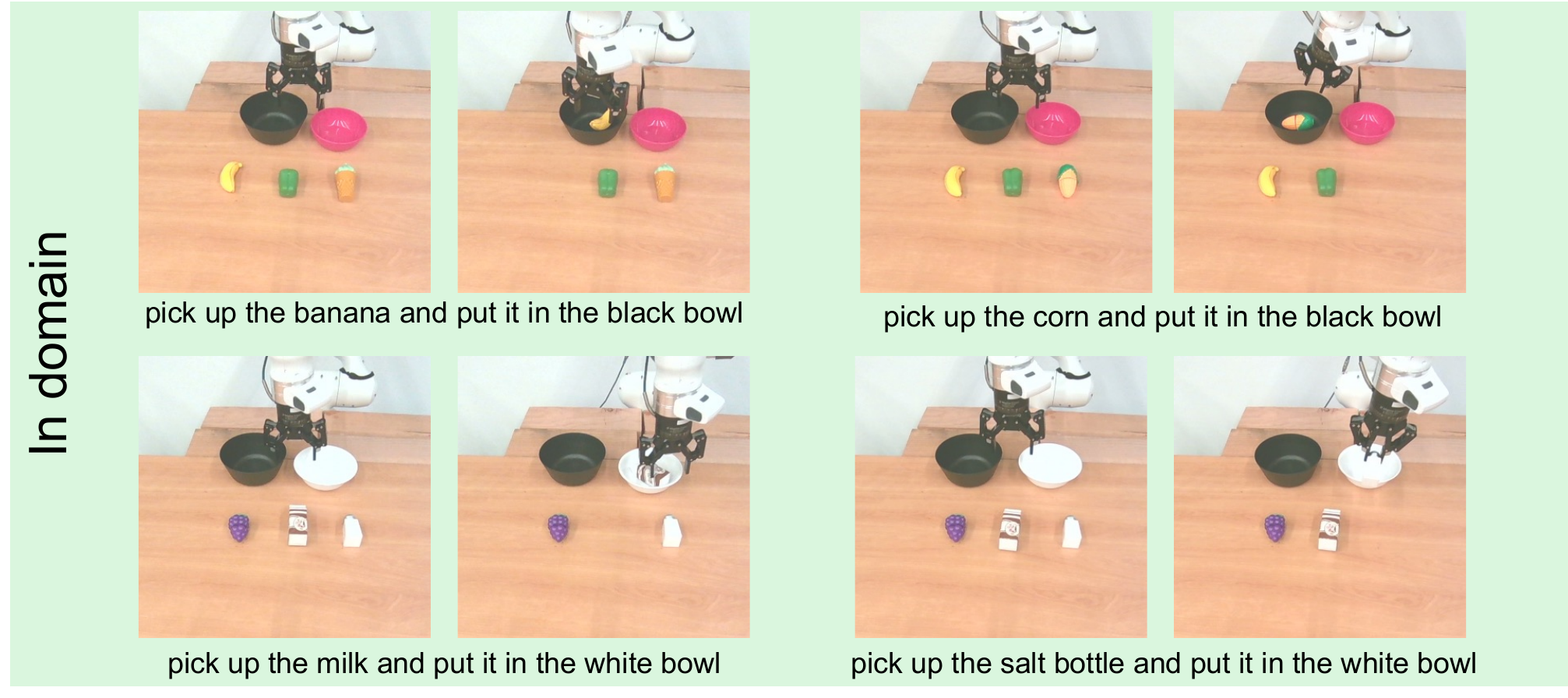}
    \caption{Illustrations of real-world tasks that we evaluated for \textit{in-domain capabilities}, where we report the detailed results in~\tabref{tab:real_world_robotics}. Specifically, we demonstrate the initial and final state of \algname in handling each of the four challenging tasks detailed in the captions.}
\label{fig:real-world-case-1}
\vspace{-2em}
\end{figure*}

\begin{figure*}[htbp]
    \centering
    \includegraphics[width=0.95\textwidth]{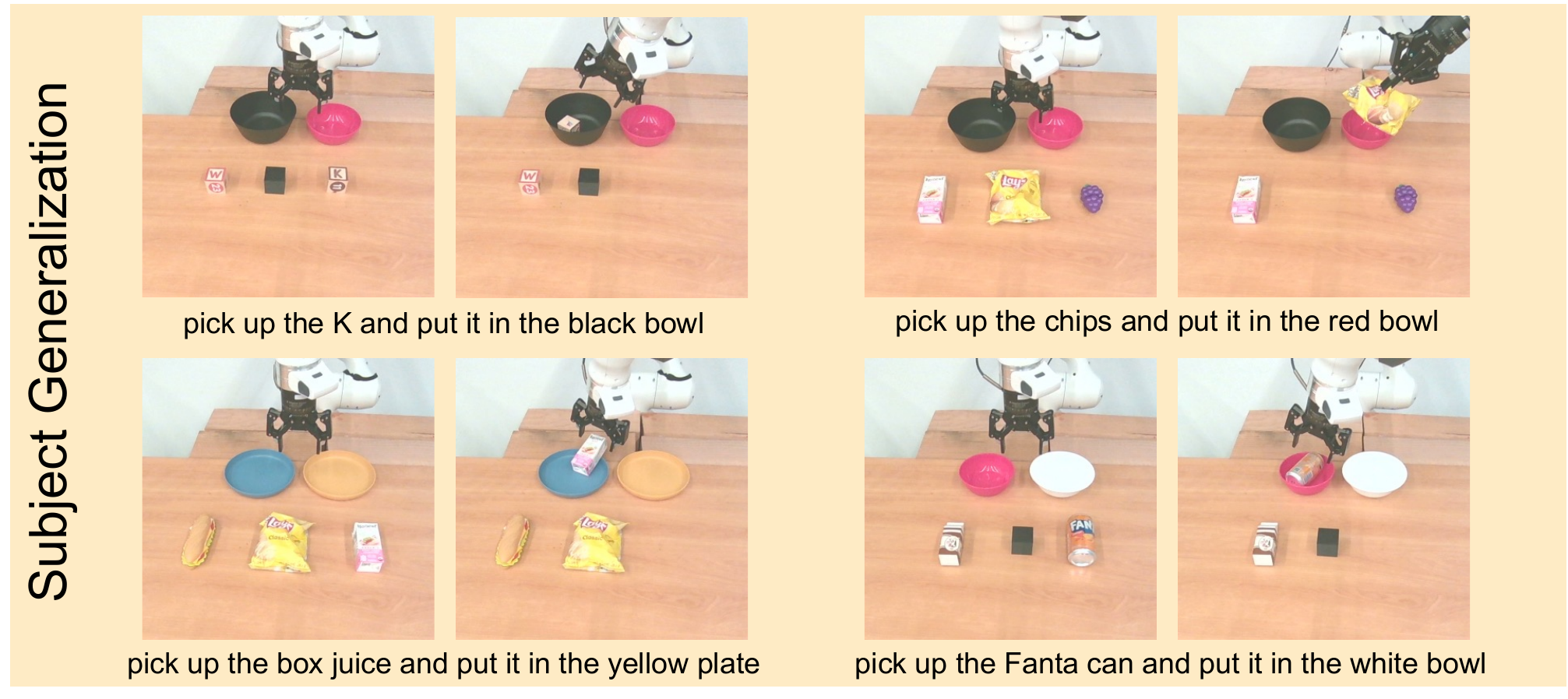}
    \caption{Illustrations of real-world tasks that we evaluated for \textit{subject generation}, where we report the detailed results in~\tabref{tab:real_world_robotics}. Specifically, we demonstrate the initial and final state of \algname in handling each of the four challenging tasks detailed in the captions.}
\label{fig:real-world-case-2}
\vspace{-2em}
\end{figure*}

\begin{figure*}[htbp]
    \centering
    \includegraphics[width=0.95\textwidth]{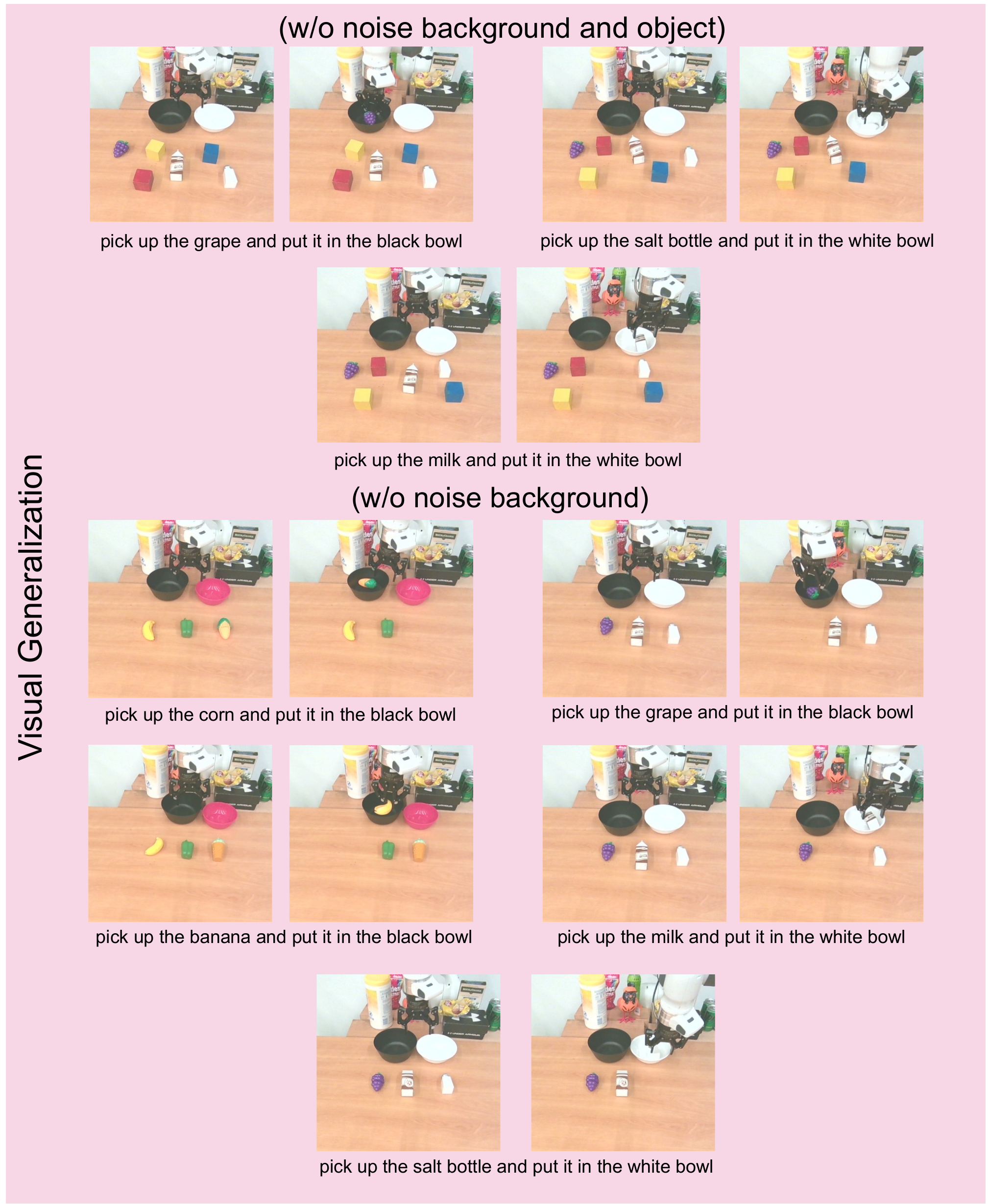}
    \caption{Illustrations of real-world tasks that we evaluated for \textit{visual generation}, where we report the detailed results in~\tabref{tab:real_world_robotics}. Specifically, we demonstrate the initial and final state of \algname in handling each of the eight challenging tasks detailed in the captions.}
\label{fig:real-world-case-4}
\end{figure*}

\begin{figure*}[htbp]
    \centering
    \includegraphics[width=0.95\textwidth]{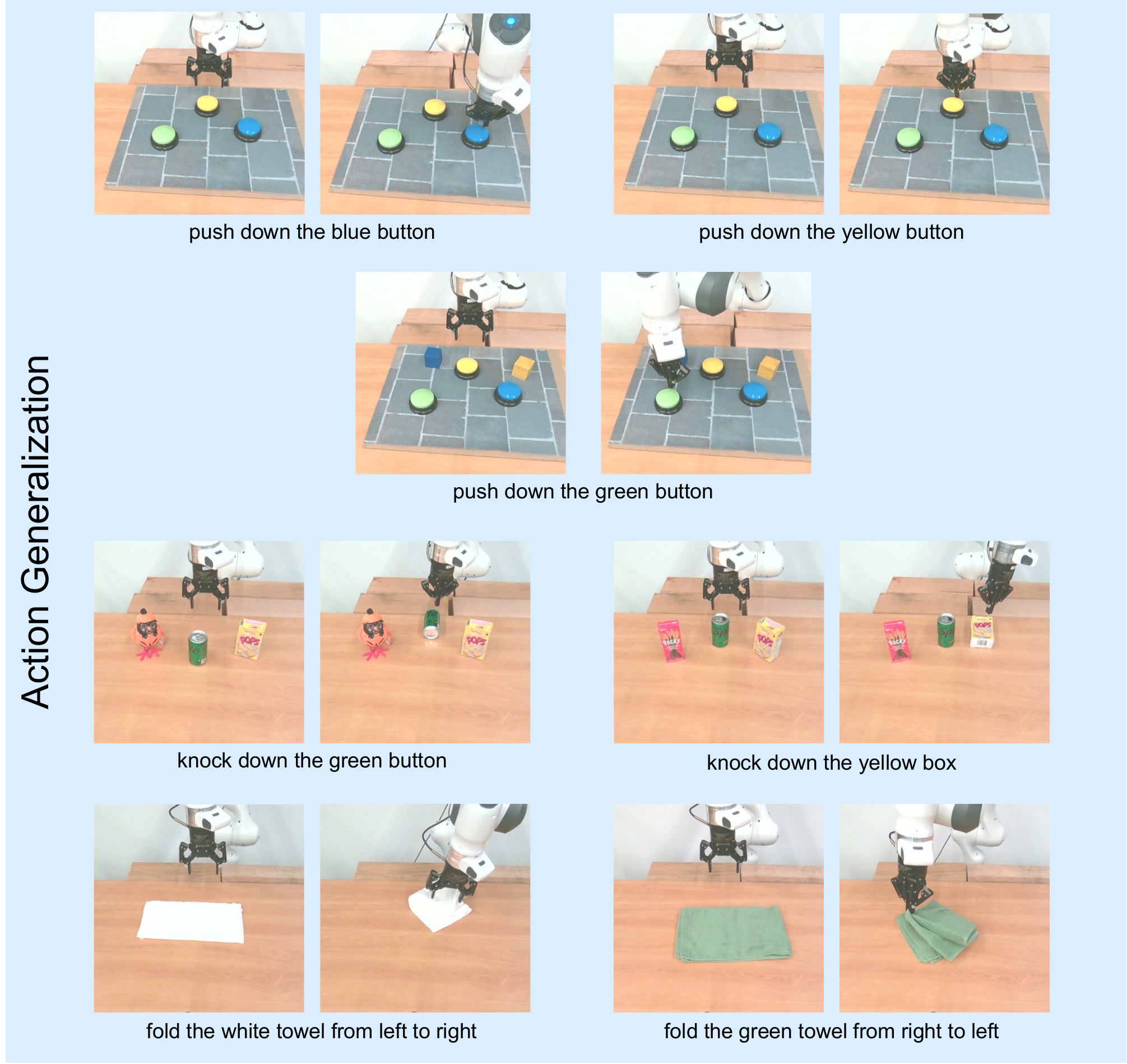}
    \caption{Illustrations of real-world tasks that we evaluated for \textit{action generation}, where we report the detailed results in~\tabref{tab:real_world_robotics}. Specifically, we demonstrate the initial and final state of \algname in handling each of the seven challenging tasks detailed in the captions.}
\label{fig:real-world-case-5}
\vspace{-2em}
\end{figure*}

\begin{figure*}[htbp]
    \centering
    \includegraphics[width=0.95\textwidth]{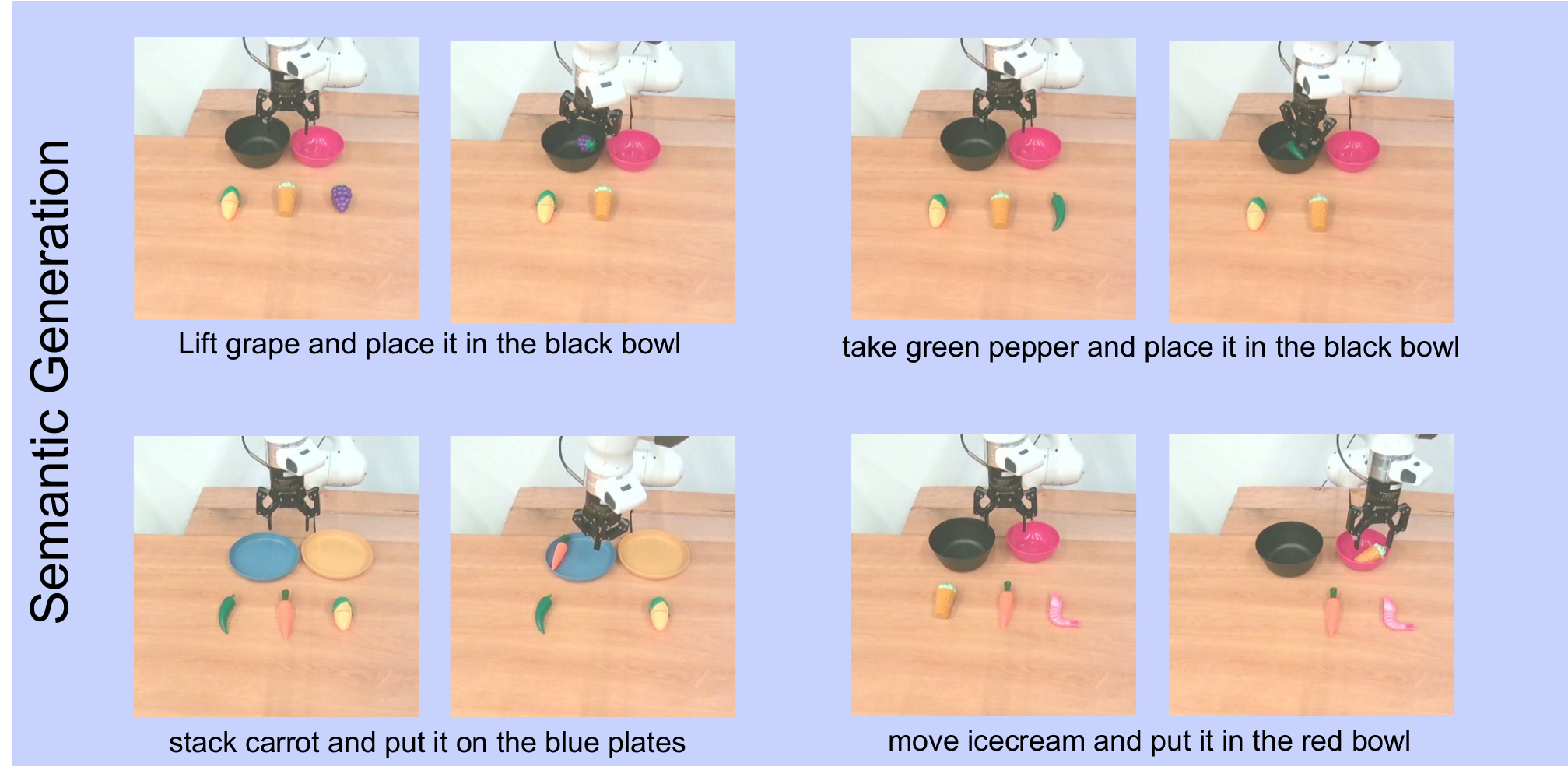}
    \caption{Illustrations of real-world tasks that we evaluated for \textit{semantic generation}, where we report the detailed results in~\tabref{tab:real_world_robotics}. Specifically, we demonstrate the initial and final state of \algname in handling each of the four challenging tasks detailed in the captions.}
\label{fig:real-world-case-6}
\end{figure*}

\begin{figure*}[htbp]
    \centering
    \includegraphics[width=0.95\textwidth]{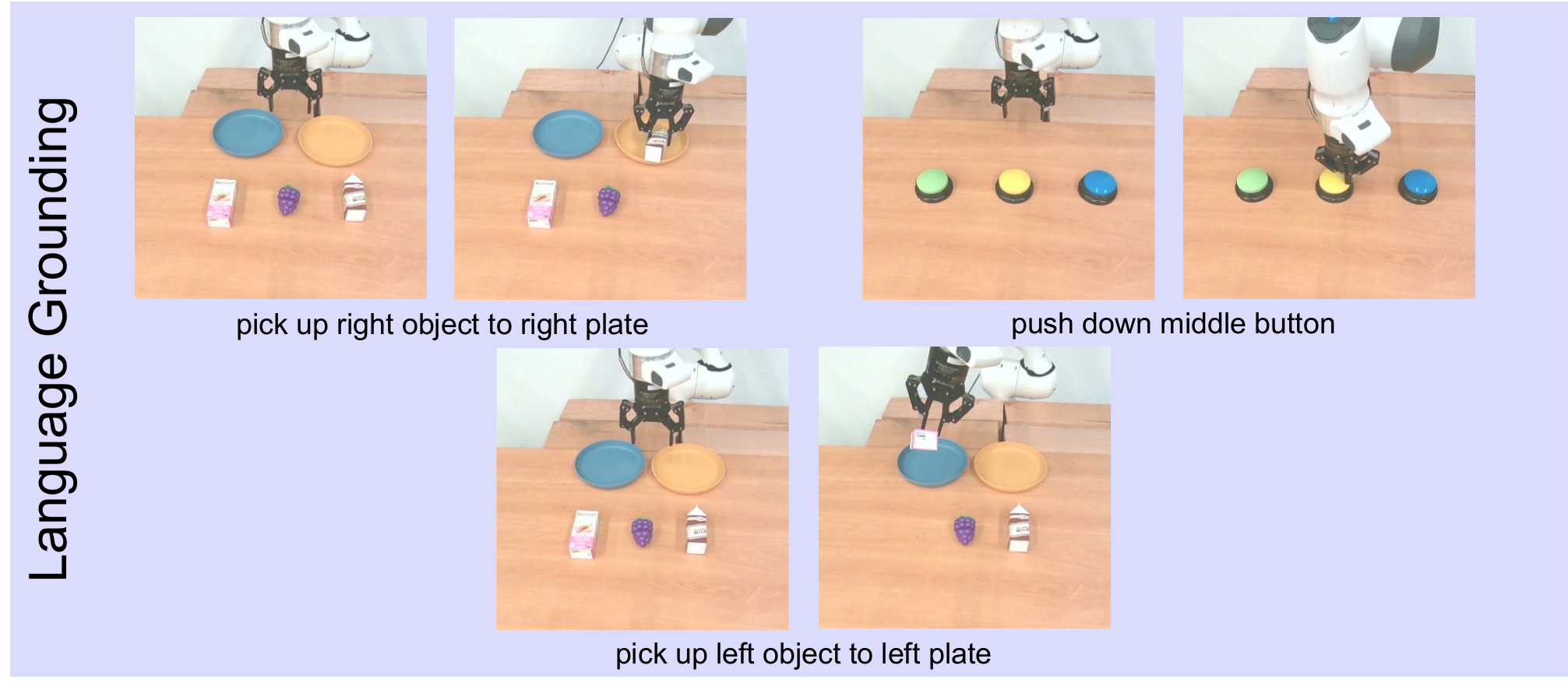}
    \caption{Illustrations of real-world tasks that we evaluated for \textit{language generation}, where we report the detailed results in~\tabref{tab:real_world_robotics}. Specifically, we demonstrate the initial and final state of \algname in handling each of the five challenging tasks detailed in the captions.}
\label{fig:real-world-case-3}
\vspace{-2em}
\end{figure*}

We provide an illustration for each specific task included in the suite evaluation for \textit{in-domain} tasks in~\figref{fig:real-world-case-1} and for each type of generation task, including \textit{subject generalization} in~\figref{fig:real-world-case-2}, \textit{language grounding} in~\figref{fig:real-world-case-3}, \textit{visual generalization} in~\figref{fig:real-world-case-4}, \textit{action generalization} in~\figref{fig:real-world-case-5}, and \textit{semantic generalization} in~\figref{fig:real-world-case-6}. Specifically, we demonstrate the initial and final states of \algname in handling each of these challenging tasks, as detailed in the corresponding captions. \revision{In addition, we include a safety task to demonstrate how GRAPE adheres to safety requirements once aligned with safety constrains.}

\begin{figure*}[htbp]
    \centering
    \includegraphics[width=0.95\textwidth]{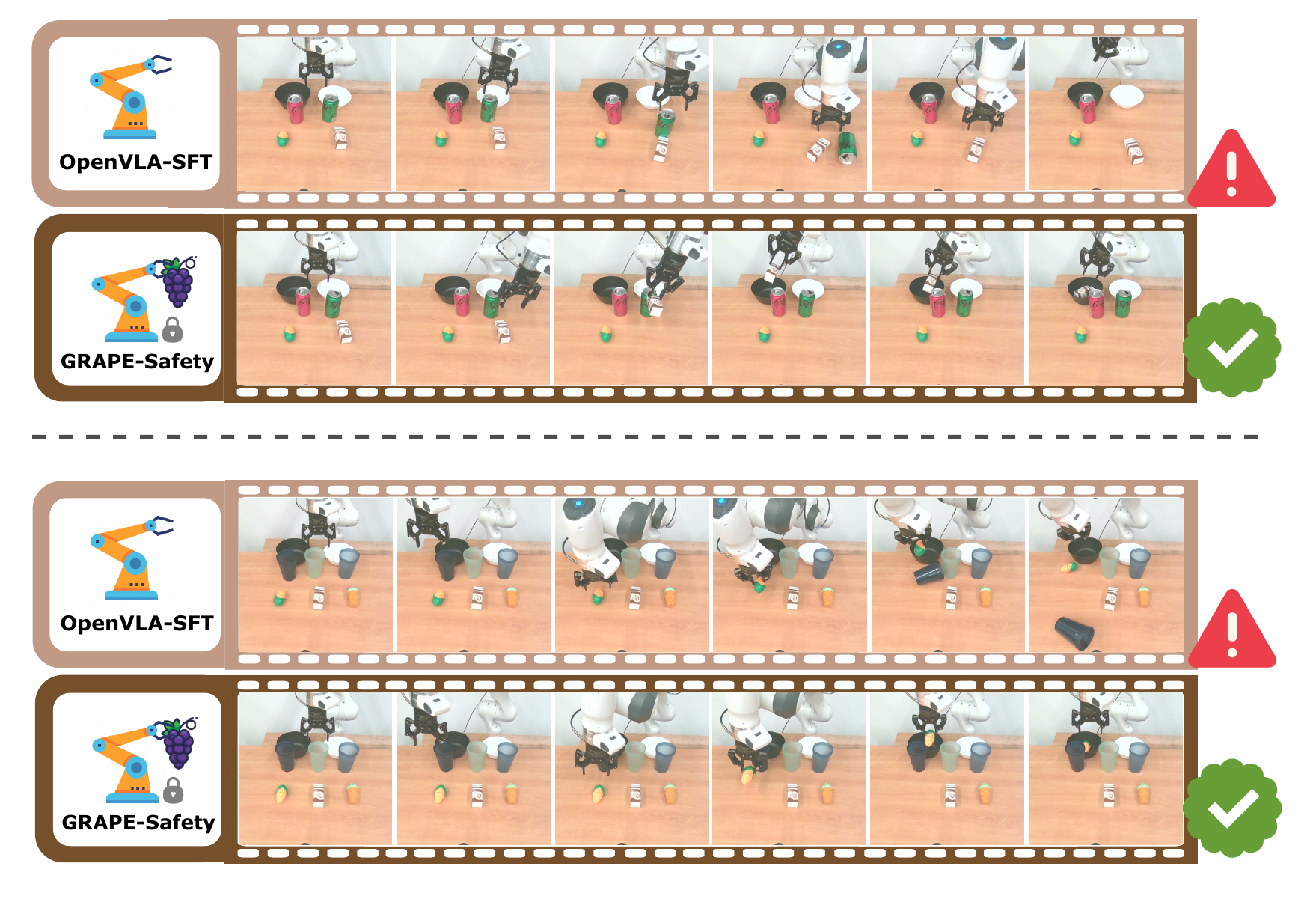}
    \caption{\revision{Illustrations of real-world tasks used for safety evaluation, extending the tasks presented in Figure 7. The figure shows key frames from GRAPE’s trajectory in two challenging scenarios. Due to the lack of safety reward alignment, the OpenVLA-SFT approach fails, while GRAPE-Safety successfully navigates obstacles and completes the task once the safety rewards are properly aligned.}}
\label{fig:real-world-case-safety}
\end{figure*}

\onecolumn
\clearpage
\subsection{Case Study of Multi-stage Cost Functions}
\label{sec:case_study_cost}
We demonstrate some case studies of the multi-stage cost functions generated using our proposed pipeline given different alignment objectives.

\subsubsection{Task Completion}
\begin{boxfour}\label{box:task_sccuess}
    \begin{verbatim}
    # The task involves picking up the grape and placing it in the black bowl.
    # The stages involved are:
    # 1. Grasp grape
    # 2. Move grape to black bowl
    # 3. Drop grape in black bowl

    num_stages = 3

    ### stage 1: Grasp grape
    
    def stage1_target_constraint1(end_effector, keypoints):
        """Align the end-effector with the grape's center."""
        
        grape_center = keypoints[0]
        target_cost = np.linalg.norm(end_effector - grape_center)
        return target_cost

    ### stage 2: Move grape to black bowl

    def stage2_target_constraint1(end_effector, keypoints):
        """Calculate the relative distance between grape and black bowl."""
        
        black_bowl_center = keypoints[1]# Assuming keypoint 1 is the black bowl
        target_cost = np.linalg.norm(end_effector - black_bowl_center)
        return target_cost

    ### stage 3: Drop grape in black bowl
    
    def stage3_target_constraint1(end_effector, keypoints):
        """Ensure the grape rests in the black bowl."""
        
        black_bowl_center = keypoints[1]
        target_cost = np.linalg.norm(end_effector - black_bowl_center)
        return target_cost    
    \end{verbatim}
\end{boxfour}
\subsubsection{Safety}
\begin{boxtwo}\label{box:safety}
    \begin{verbatim}
    # The task involves picking up the grape and placing it in the black bowl.
    # The stages involved are:
    # 1. Grasp grape
    # 2. Move grape to black bowl
    # 3. Drop grape in black bowl

    num_stages = 3

    ### stage 1: Grasp grape
    
    def stage1_collision_constraint1(end_effector, keypoints):
        """Approach the grape from above to avoid collision."""
        
        grape_center = keypoints[0]
        collision_cost = 0 if end_effector[1] > grape_center[1] else 1
        return collision_cost

    ### stage 2: Move grape to black bowl

    def stage2_collision_constraint1(end_effector, keypoints):
        """Ensure the grape is aligned above the black bowl."""
        
        obstacles = keypoints[2:]#Assuming keypoints[2:] are obstacles
        threshold = 0.1  # Minimum allowable clearance
        collision_cost = sum(
            max(0, threshold - np.linalg.norm(end_effector - obstacle))
            for obstacle in obstacles
        )
        return collision_cost

    ### stage 3: Drop grape in black bowl
    
    def stage3_collision_constraint1(end_effector, keypoints):
        """Approach the grape from above to avoid collision."""
        
        black_bowl_center = keypoints[1]
        collision_cost = 0 if end_effector[1] > black_bowl_center[1] else 1
        return collision_cost
    \end{verbatim}
\end{boxtwo}

\subsubsection{Cost-Efficiency}

\begin{boxthree}\label{box:efficiency}
    \begin{verbatim}
    # The task involves picking up the grape and placing it in the black bowl.
    # The stages involved are:
    # 1. Grasp grape
    # 2. Move grape to black bowl
    # 3. Drop grape in black bowl

    num_stages = 3

    ### stage 1: Grasp grape
    
    def stage1_path_constraint1(end_effector, keypoints):
        """Align the end-effector with the grape's center."""
        
        grape_center = keypoints[0]
        distance = np.linalg.norm(end_effector - grape_center)
        step_size = 0.01  # Assuming a small step size
        path_cost = int(distance / step_size)
        return path_cost

    ### stage 2: Move grape to black bowl

    def stage2_path_constraint1(end_effector, keypoints):
        """Calculate the relative distance between grape and black bowl."""
        
        black_bowl_center = keypoints[1]# Assuming keypoint 1 is the black bowl
        distance = np.linalg.norm(end_effector - black_bowl_center)
        step_size = 0.01  # Assuming a small step size
        path_cost = int(distance / step_size)
        return path_cost

    ### stage 3: Drop grape in black bowl
    
    def stage3_path_constraint1(end_effector, keypoints):
        """Ensure the grape rests in the black bowl."""
        
        black_bowl_center = keypoints[1]
        distance = np.linalg.norm(end_effector - black_bowl_center)
        step_size = 0.01  # Assuming a small step size
        path_cost = int(distance / step_size)
        return path_cost    
    \end{verbatim}
\end{boxthree}

\clearpage
\begin{boxone}\label{box:prompt}
\begin{dialogue}
\speak{\textbf{User}} 
    \subsection*{Instructions}
    
    The image shows a robot stage point in a workspace, each point in the diagram represents the point of the stage split:
    \begin{itemize}
         \item Stage\_point\_0 : Represents the initial position of the carrot.
         \item Stage\_point\_1 : Represents the intermediate position above the carrot for grasping.
    \end{itemize}

    Determine how many stages are involved in the task. Grasping must be an independent stage. Some examples:

    \subsubsection*{1. Task: Put the carrot on the plate}
    
    \paragraph{Stages:}
    
    \begin{itemize}
        \item \textbf{Grasp carrot}
        \item \textbf{Move carrot to plate}
        \item \textbf{Drop carrot on plate}
    \end{itemize}
    
    \paragraph{Stage 1: Grasp carrot}
    
    \begin{itemize}
        \item \textbf{Path constraints}:
        \begin{itemize}
            \item Align the end-effector with the carrot's center.
        \end{itemize}
        \item \textbf{Collision constraints}:
        \begin{itemize}
            \item The end-effector must approach the carrot from above to avoid collision.
        \end{itemize}
    \end{itemize}
    
    \paragraph{Stage 2: Move carrot to plate}
    
    \begin{itemize}
        \item \textbf{Path constraints}:
        \begin{itemize}
            \item Calculate the relative distance between carrot and plate.
        \end{itemize}
        \item \textbf{Collision constraints}:
        \begin{itemize}
            \item The carrot is aligned above the plate.
        \end{itemize}
    \end{itemize}
    
    \paragraph{Stage 3: Drop carrot on plate}
    
    \begin{itemize}
        \item \textbf{Path constraints}:
        \begin{itemize}
            \item The carrot must rest on the plate.
            \item The carrot should not bounce out of the basket.
        \end{itemize}
        \item \textbf{Collision constraints}:
        \begin{itemize}
            \item The end-effector must approach the carrot from above to avoid collision.
        \end{itemize}
    \end{itemize}
    
    \bigskip
    
    \textbf{Note:}
    
    \begin{itemize}
        \item Sum all Path constraints cost the \verb|path_cost| variable.
        \item Sum all Grasp constraints cost the \verb|grasp_cost| variable.
        \item Sum all Collision constraints cost the \verb|collision_cost| variable.
        \item Each constraint function takes an end-effector point and a set of keypoints as input, returning a numerical cost. The constraint is satisfied if this cost is zero or less.
        \item Define any number of path constraints per stage, but avoid using "if" statements in the functions.
        \item Avoid using path constraints when manipulating deformable objects (e.g., towels).
        \item Input format:
        \begin{itemize}
            \item \verb|end_effector|: \verb|np.array| of shape \verb|(3,)| representing the end-effector position.
            \item \verb|keypoints|: \verb|np.array| of shape \verb|(K, 3)| representing the keypoints positions.
        \end{itemize}
        \item Use Python and NumPy functions freely in constraint functions.
        \item Use pairs of keypoints to create vectors if needed.
        \item Keypoints are indexed starting from 0, matching their order in the keypoints array.
    \end{itemize}
    
    \bigskip
    
    \textbf{Structure your output in a single Python code block as follows:}
    
    \begin{verbatim}
    # ...
    
    num_stages = ?
    
    ### stage 1 path constraints (if any)
    def stage1_path_constraint1(end_effector, keypoints):
        """Put your explanation here."""
        ...
        return path_cost
    # Add more constraints if needed
    ...
    
    ### stage 1 collision constraints (if any)
    def stage1_collision_constraint1(end_effector, keypoints):
        """Put your explanation here."""
        ...
        return collision_cost
    
    # Add more constraints if needed
    ...
    
    # Repeat for more stages
    ...
    \end{verbatim}

\bigskip

\textbf{Query}

Query Task: "\{instruction\}"

Query Image:

\end{dialogue}
\end{boxone}
\twocolumn  




\end{document}